\documentclass[10pt,twocolumn,letterpaper]{article}
\usepackage{standalone}
\usepackage{titling}
\usepackage{iccv}
\usepackage{times}
\usepackage{epsfig}
\usepackage{graphicx}
\usepackage{amsmath}
\usepackage{amssymb}
\usepackage{mathrsfs}
\usepackage{multirow}

\usepackage{subfigure}

\usepackage{tabularx}
\usepackage[accsupp]{axessibility}

\usepackage[toc,page,titletoc]{appendix}

\usepackage[breaklinks=true,bookmarks=false]{hyperref}

\iccvfinalcopy 

\begin{document}


\title{Unsupervised Dense Deformation Embedding Network for \\ Template-Free Shape Correspondence}
\date{}
\author{Ronghan Chen\textsuperscript{1,2,3}
\and
Yang Cong\textsuperscript{1,2}\thanks{The corresponding author is Prof. Yang Cong.}
\and
Jiahua Dong\textsuperscript{1,2,3}
\and
\textsuperscript{1}State Key Laboratory of Robotics, Shenyang Institute of Automation, Chinese Academy of Sciences\thanks{This work is supported in part by the National Key Research and Development Program of China under Grant 2019YFB1310300 and the National Nature Science Foundation of China under Grant 61821005.}\\
\textsuperscript{2}Institutes for Robotics and Intelligent Manufacturing, Chinese Academy of Sciences\\
\textsuperscript{3}University of Chinese Academy of Sciences\\
{\tt\small chenronghan@sia.cn, congyang81@gmail.com, dongjiahua1995@gmail.com}
}

\maketitle


\begin{abstract}

Shape correspondence from 3D deformation learning has attracted appealing academy interests recently. Nevertheless, current deep learning based methods require the supervision of dense annotations to learn per-point translations, which severely over-parameterize the deformation process. Moreover, they fail to capture local geometric details of original shape via global feature embedding. To address these challenges, we develop a new \underline{U}nsupervised \underline{D}ense \underline{D}eformation \underline{E}mbedding \underline{Net}work (\emph{i.e.}, UD$^2$E-Net), which learns to predict deformations between non-rigid shapes from dense local features. Since it is non-trivial to match deformation-variant local features for deformation prediction, we develop an Extrinsic-Intrinsic Autoencoder to first encode extrinsic geometric features from source into intrinsic coordinates in a shared canonical shape, with which the decoder then synthesizes corresponding target features. Moreover, a bounded maximum mean discrepancy loss is developed to mitigate the distribution divergence between the synthesized and original features. To learn natural deformation without dense supervision, we introduce a coarse parameterized deformation graph, for which a novel trace and propagation algorithm is proposed to improve both the quality and efficiency of the deformation. Our UD$^2$E-Net outperforms state-of-the-art unsupervised methods by 24$\%$ on Faust Inter challenge and even supervised methods by 13$\%$ on Faust Intra challenge.
	
\end{abstract}

\begin{figure}[t]
	\centering
	\includegraphics[width=0.48\textwidth]{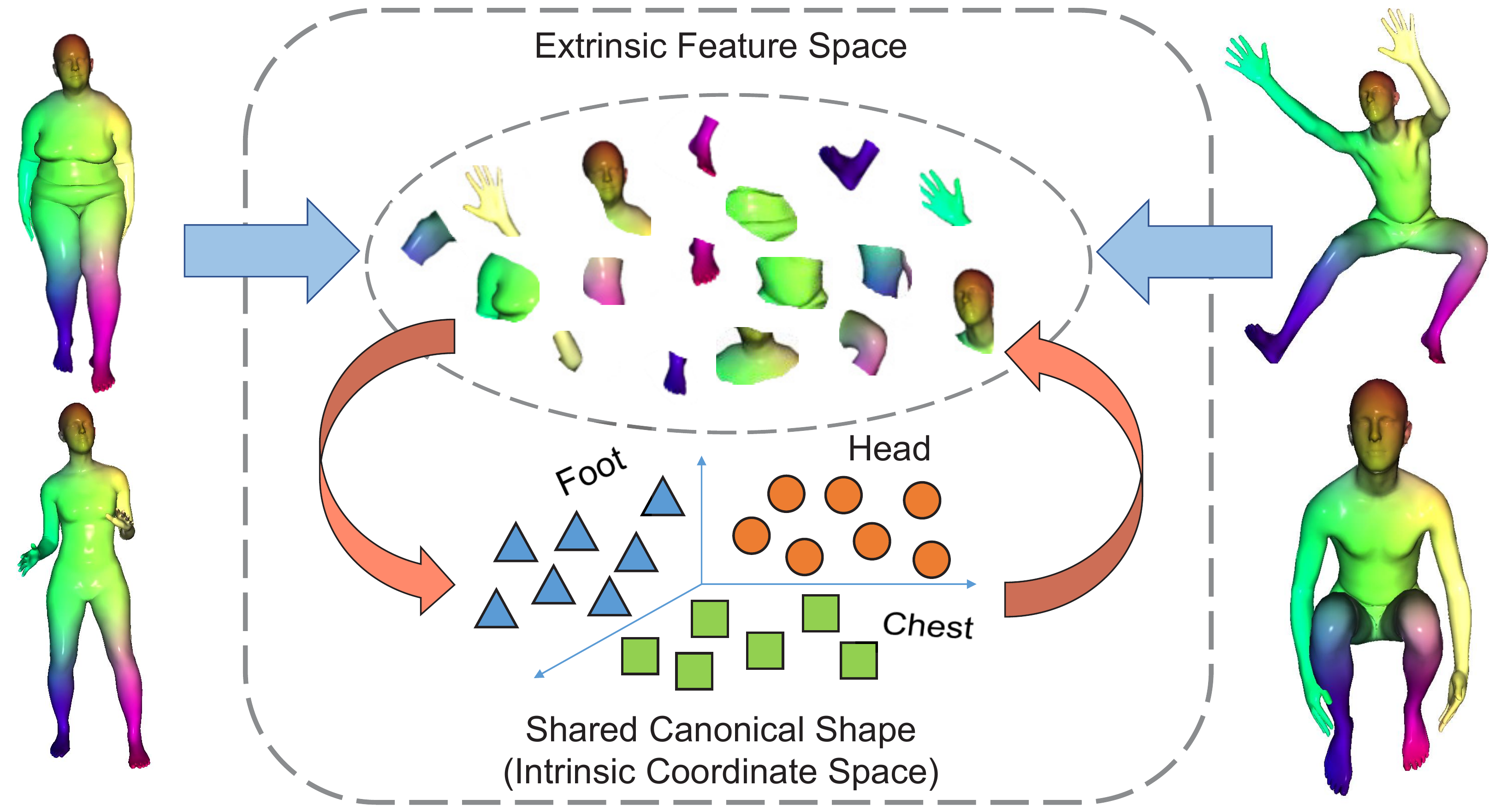}
	\caption{Illustration of our UD$^2$E-Net model, which achieves the bidirectional mapping between extrinsic feature space and intrinsic coordinate space for dense deformation embedding.}\label{fig:intro}
\end{figure}
\section{Introduction}
Alignment of deformable 3D shapes is a ubiquitous challenge in the field of computer vision and graphics with many applications, including non-rigid reconstruction, deformation transfer and texture mapping. Traditional methods~\cite{sumner2007embedded, li2008global} infer parametric deformation by optimizing correspondences construction and objective minimization iteratively. 
However, the optimization heavily depends on initialization and is prone to being stuck in local minima, especially for large articulated deformations. With the advent of 3D deep learning techniques~\cite{pointnet, pointnet2}, deep deformation learning methods~\cite{3d-coded, atlasnet,3dn,foldingnet, li2019lbs} leveraging a large amount of data have been applied for large deformation prediction. Generally, these techniques directly regress dense translations or positions for all input 3D points, which neglects the implicit deformation disciplines and severely over-parameterizes the deformation leading to high-frequency artifacts. Unfortunately, these approaches require strong supervision of dense correspondences (\emph{e.g.}, usually more than 5000 per human body shape), which consumes huge labor-intensive efforts.  

Furthermore, most existing state-of-the-art methods~\cite{li2019lbs, 3d-coded, atlasnet, deprelle2019learning, 3dn, NeuralCage, cycle-consistent} encode the target shape into a global feature for deformation learning, which neglects the low-level geometric details and fails to infer fine-grained deformation. To leverage local features for deformation prediction, connections between both sides should be established to enable feature communication, which is a challenging correspondence problem for deformable shapes.

To address these challenges, a new \underline{U}nsupervised \underline{D}ense \underline{D}eformation \underline{E}mbedding \underline{Net}work (\emph{i.e.}, UD$^2$E-Net) endowed with the traditional \emph{Embedded Deformation} (ED) technique~\cite{sumner2007embedded} is designed to predict deformation between arbitrary source and target shape pairs. With the local rigidity regularization provided by ED, UD$^2$E-Net can learn a more natural deformation space, which we expect to mitigate the strong reliance of deep learning models on abundant annotated data.
Moreover, our network employs dense local feature embedding and fusion to reason about deformation parameters for each node within the deformation graph constructed via ED. Specifically, with the fine-grained geometric features extracted via the Siamese mesh encoder, an \emph{Extrinsic-Intrinsic Autoencoder} (EI-AE) is developed to first encode source features into intrinsic coordinates of a shared canonical shape. With the coordinates, corresponding target features are synthesized via the decoder.
To minimize the distribution gap between the synthesized and original target features, we design a Bounded Maximum Mean Discrepancy loss, which further provides the self-supervised guidance for canonical shape construction. To eliminate artifacts caused by wrong adjacent relationships between the input shape and its deformation graph via Euclidean measure, a trace and propagation algorithm is developed to improve both efficiency and accuracy by leveraging a pre-constructed mesh hierarchy. Extensive experiments demonstrate that UD$^2$E-Net shows stable performance under data volume reduction. On Faust benchmark~\cite{FAUST}, the proposed UD$^2$E-Net outperforms state-of-the-art unsupervised methods by 24$\%$$\sim$37$\%$ on Inter challenge, and shows 13$\%$ improvement on Intra challenge over supervised methods. Furthermore, experiments also demonstrate the potential of our UD$^2$E-Net in several challenging applications, \emph{e.g.}, shape retrieval and human pose transfer.

In conclusion, the main contributions of our work are:

\begin{itemize}
	\setlength{\itemsep}{5pt}		
	\setlength{\parsep}{0pt}	
	\setlength{\parskip}{0pt}
	
	\item A new {U}nsupervised {D}ense {D}eformation {E}mbedding {Net}work (\emph{i.e.}, UD$^2$E-Net) is developed to learn large articulated deformations between arbitrary shape pairs without supervision of ground-truth correspondences.
	
	\item 	We design an Extrinsic-Intrinsic Autoencoder to encode extrinsic geometric features from source shape into a shared canonical shape, which is utilized to decode corresponding synthesized target features. Meanwhile, a bounded maximum mean discrepancy loss is introduced to mitigate the distribution divergence between the synthesized and original target features.

	\item  A trace and propagation algorithm is developed to avoid artifacts brought by incorrect node-to-vertex assignment in Embedded Deformation, which improves both quality and efficiency of the deformation process.
	
\end{itemize}

\section{Related Work}
\noindent
\textbf{Non-rigid Shape Matching.}
Generally, methods for shape correspondence can be divided into intrinsic methods and extrinsic methods. Intrinsic methods rely on intrinsic property of shapes, such as spectral~\cite{hks,bronstein2010scale} or learning based~\cite{wang2018learning} descriptors or preservation of geodesic distances~\cite{FMNet} to directly obtain correspondences. However, most intrinsic methods are not robust enough under perturbation, such as topology changes and incompleteness. Other extrinsic methods achieve correspondences by explicitly deforming one shape to align the other~\cite{li2009robust, li2008global}, where hand-crafted deformation models~\cite{sumner2007embedded,sorkine2007rigid} are applied for regularization. 
However, they are prone to being stuck in local minima.

\noindent
\textbf{Deformation Representation for Learning.}
A direct and most commonly applied way to deform a 3D model is to assign each point a vector, representing translation or position~\cite{atlasnet, 3d-coded, 3dn, deprelle2019learning}. Such deformation is endued with high degree of freedom, which binds the number of predicted parameters to resolution of the input model. For low-dimensional deformation representations, free-form deformation~\cite{ffd} is utilized in~\cite{jack2018learning} to deform meshes by manipulating several predefined vertices of structured grids. LBS autoencoder~\cite{li2019lbs} uses predefined LBS model to deform human body and hands, which cannot adapt to general objects. \cite{NeuralCage} utilizes the cage-based deformation technique~\cite{ju2005mean} for detail-preserving deformation and cannot achieve perfect alignment. DEMEA~\cite{demea} proposes an embedded deformation layer based on ED~\cite{sumner2007embedded} to learn mesh representations. Obviously, ED is vastly limited by fully connected layers, which are known to be permutation variant and cannot deal with unordered point clouds and meshes. 

\noindent
\textbf{Deep Deformation Embedding.}
Existing methods typically embed the target shape into a global feature, and then rely on a vector it concatenates with for location cues to express different deformation information~\cite{atlasnet, 3d-coded, deprelle2019learning, NeuralCage, 3dn, Completion}. However, they cannot provide fine geometric details. Other methods infer deformation from local features. Pixel2Mesh~\cite{pixel2mesh, wen2019pixel2mesh++} assigns local features of 2D image to vertices of an ellipsoid template by 3D-2D projection. Flownet3d~\cite{flownet3d} mixes features from two point clouds based on Euclidean neighbours, which are not reliable under large articulated deformations. Unlike them, our model can learn stable feature pairs via the canonical shape built by EI-AE.

\section{Our Proposed Model}

\begin{figure*}[htbp]
	\centering
	\includegraphics[width=1.\linewidth]{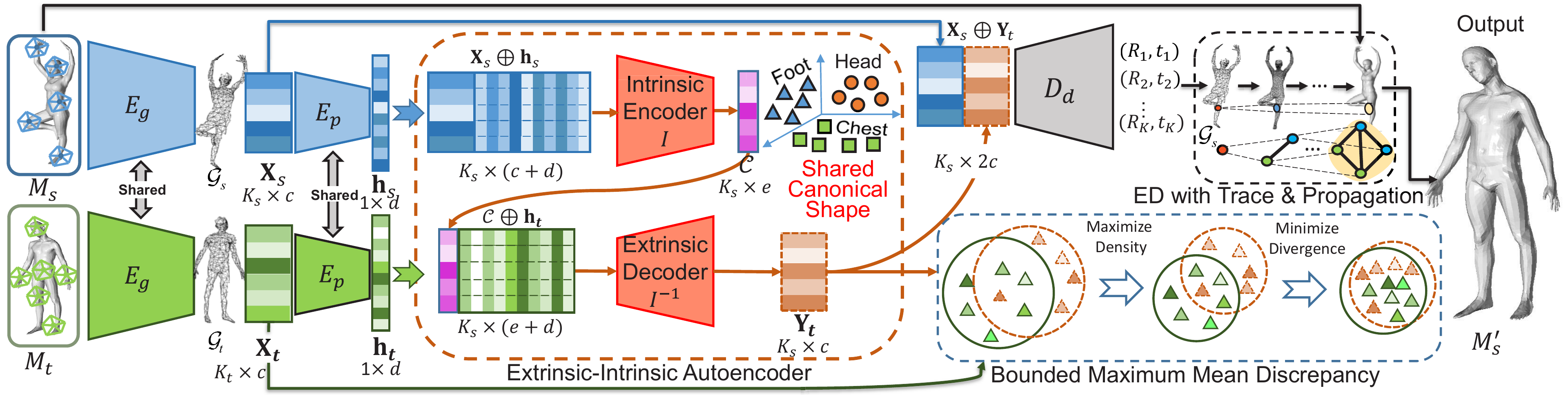}
	\caption{Overview architecture of UD$^2$E-Net.
		After extracting local features $\mathbf{X}_s,\mathbf{X}_t$ via the Siamese mesh encoder $E_g$ and global features $\mathbf{h}_s,\mathbf{h}_t$ via a PointNet-like network $E_p$, the intrinsic encoder $\mathcal{I}$ in EI-AE encodes $\mathbf{X}_s$ into the coordinates in the shared canonical shape $\mathcal{C}$, which is concatenated with $\mathbf{h}_t$ and further decoded by the extrinsic decoder $\mathcal{I}^{-1}$ to obtain corresponding local features $\mathbf{Y}_t$. The bounded maximum mean discrepancy is applied to eliminate the domain gap between $\textbf{X}_t$ and $\mathbf{Y}_t$. Then the deformation decoder $D_d$ decodes the concatenation of $\mathbf{Y}_t$ and $\mathbf{X}_s$ into deformation parameters in $\mathcal{G}_s$ to output the deformation prediction of $M_s$.} \label{fig:main}
	\vspace{-5pt}
\end{figure*}

\subsection{Preliminary}\label{deformation_graph}

\paragraph{Embedded Deformation.} We build upon the \emph{Embedded Deformation} (ED) algorithm~\cite{sumner2007embedded} to model non-rigid deformation. To deform a mesh $M=(V, E)$ with $N$ vertices $V=\{\boldsymbol{v}_i\}_{i=1}^N$ and edges $E$, ED constructs a coarse embedded deformation graph $\mathcal{G}$ to manipulate $M$. $\mathcal{G}=(\mathcal{N},\mathcal{E})$ is a low-resolution $M$ and is composed of a node set $\mathcal{N}=\{\boldsymbol{g}_i\}_{i=1}^K$ and an edge set $\mathcal{E}$. Each node $\boldsymbol{g}_{i}\in\mathbb{R}^3$ is parameterized with transformation $\boldsymbol{T}_i=(\boldsymbol{R}_i,\boldsymbol{t}_i)$ to control the local deformation, where $\boldsymbol{R}_i$ is a rotation matrix and $\boldsymbol{t}_i\in\mathbb{R}^3$ is a translation vector. With local deformations $T=\{(\boldsymbol{R}_i,\boldsymbol{t}_i)\}_{i=1}^K$, $M$ can be deformed by:
\begin{equation}\label{equ:deform}
\boldsymbol{v}_j'=\sum_{\boldsymbol{g}_i\in \mathcal{A}(\boldsymbol{v}_j)}w_{i,j}(\boldsymbol{R}_i(\boldsymbol{v}_j-\boldsymbol{g}_i)+\boldsymbol{g}_i+\boldsymbol{t}_i),
\end{equation}
where $\boldsymbol{v}_j$ is the position of a vertex on $M$ and $\boldsymbol{v}_j'$ is the deformed $\boldsymbol{v}_j$. $\mathcal{A}(\boldsymbol{v}_j)$ is the set of nodes controlling the deformation of vertex $\boldsymbol{v}_j$, and $w_{i,j}$ is the node-to-vertex weight determined by distance between $\boldsymbol{v}_j$ and $\boldsymbol{g}_i$. 
Finally, an as-rigid-as-possible loss is utilized to further balance the deformation and smoothness: 
\begin{equation}\label{arap}
\mathcal{L}_{\rm ARAP}=\sum_{i}\sum_{j:(i,j)\in\mathcal{E}}\|(\boldsymbol{R}_i(\boldsymbol{g}_j-\boldsymbol{g}_i)+\boldsymbol{g}_i+\boldsymbol{t}_i)-(\boldsymbol{g}_j+\boldsymbol{t}_j)\|,
\end{equation}
where $\boldsymbol{g}_i,\boldsymbol{g}_j$ are adjacent nodes connected with an edge in $\mathcal{E}$. The $\mathcal{L}_{\rm ARAP}$ term indicates that the local deformations of adjacent nodes should agree with one another.

Given a source mesh $M_s$ and a target mesh $M_t$, as shown in Figure~\ref{fig:main}, the goal of our unsupervised dense deformation embedding network (UD$^2$E-Net) is to output a deformed $M_s'$ to align with $M_t$ for matching, by predicting the local transformation $T=\{(\boldsymbol{R}_i, \boldsymbol{t}_i)\}_{i=1}^K$ of the deformation graph $\mathcal{G}$. To learn these transformations, we propose a novel Extrinsic-Intrinsic Autoencoder to correlate the extracted local geometric information from input shapes and fuse them into dense deformation embeddings, which is introduced as follows.

\subsection{Extrinsic-Intrinsic Autoencoder}

As shown in Figure~\ref{fig:main}, we utilize a Siamese mesh encoder $\mathbf{E}_g$ to encode source shape $M_s$ and target shape $M_t$ into dense geometric embeddings for each node in deformation graph.
Due to incompatible resolution between input meshes and deformation graphs, we utilize graph pooling layers to gradually downsample $M_s$ and $M_t$ into deformation graphs $\mathcal{G}_s$ and $\mathcal{G}_t$, which forms a mesh hierarchy. The outputs of Siamese mesh encoder are two sets of $c$-dimensional features $\mathbf{X}_s=[\mathbf{x}^s_{1},\cdots,\mathbf{x}^s_{K_s}]\in\mathbb{R}^{K_s\times c}$ and $\mathbf{X}_t=[\mathbf{x}^t_{1},\cdots,\mathbf{x}^t_{K_t}]\in\mathbb{R}^{K_t\times c}$, as shown in Figure~\ref{fig:main}, where each feature $\mathbf{x}^s_{i}\in\mathbb{R}^{c} (i=1,\cdots,K_s)$ corresponds to a node $\boldsymbol{g}^s_{i}$ in the node set $\mathcal{N}_s$ of $\mathcal{G}_s$ and same for $\mathbf{x}^t_{j}$. Then, we utilize a PointNet-like~\cite{pointnet} encoder $\mathbf{E}_p$ to aggregate $\mathbf{X}_s$ and $\mathbf{X}_t$ into $d$-dimensional global feature vectors $\mathbf{h}_s, \mathbf{h}_t\in\mathbb{R}^{d}$, respectively.

Typically, most previous methods~\cite{3d-coded, deprelle2019learning, 3dn} employ the global feature for deformation reasoning, which may lose local geometric information thus degrades the performance of deformation prediction. To tackle this issue, we utilize local features from source and target shapes ($\mathbf{X}_s$ and $\mathbf{X}_t$) to embed deformation information for nodes in $\mathcal{N}_s$. However, it is non-trivial to match these deformation-variant local features. Former methods leverage projection~\cite{pixel2mesh}, or find nearest points in Euclidean space~\cite{flownet3d} to directly match features from source and target shapes. Nevertheless, they can lead to faulty correspondences due to large articulated deformation between input shapes. A seemingly promising solution is to learn deformation-invariant features embedded with intrinsic property~\cite{prnet} for matching, which, however, forces them to abandon essential extrinsic geometric information for deformation prediction.

Therefore, we propose a novel \emph{Extrinsic-Intrinsic Autoencoder} (EI-AE) to learn both intrinsic properties for matching and extrinsic properties for deformation prediction. As shown in Fig.~\ref{fig:main}, it constructs a shared canonical shape $\mathcal{C}=[\mathbf{c}_1,\cdots,\mathbf{c}_{K_s}]\in\mathbb{R}^{K_s\times e}$, where $e$ is the dimension of the canonical shape and $\mathbf{c}_i$ is the intrinsic coordinate shared by corresponding vertices on all shapes. Given the local feature $\mathbf{x}^s_i$ and the global feature of source mesh $\mathbf{h}_s$, EI-AE learns to parameterize each node $\boldsymbol{g}^s_{i}$ in $\mathcal{G}_s$ with its intrinsic coordinate $\mathbf{c}_i$ via an \emph{Intrinsic Encoder} $\mathcal{I}$:
\begin{equation}
\mathcal{C}=\mathcal{I}(\mathbf{X}_s\oplus\mathbf{h}_s),
\end{equation}
where $\oplus$ represents feature concatenation operation.
Due to coarseness of $\mathcal{G}_s$ and $\mathcal{G}_t$, there are often no exact matches between $\mathbf{X}_s$ and $\mathbf{X}_t$. Thus, we utilize the intrinsic coordinates in $\mathcal{C}$ to synthesize the corresponding extrinsic target features $\mathbf{Y}_t=[\mathbf{y}^t_{1},\cdots,\mathbf{y}^t_{K_s}]\in \mathbb{R}^{K_s\times c}$ from the global feature of target shape $\mathbf{h}_t$ via the \emph{Extrinsic Decoder} $\mathcal{I}^{-1}$:
\begin{equation}
\mathbf{Y}_t=\mathcal{I}^{-1}(\mathcal{C}\oplus\mathbf{h}_t).
\end{equation}
Note that, in this way, EI-AE can provide ultimate flexibility as it can correlate features from two deformation graphs $\mathcal{G}_s$, $\mathcal{G}_t$ with different number of nodes. Since the synthesized features $\mathbf{y}^t_i$ in $\mathbf{Y}_t$ and $\mathbf{x}^s_i$ in $\mathbf{X}_s$ are expected to be corresponded, the final output of EI-AE is the concatenation of $\mathbf{X}_s$ and $\mathbf{Y}_t$, \emph{i.e.},  $[\mathbf{X}_s,\mathbf{Y}_t]\in\mathbb{R}^{K_s\times2c}$, which are then sent into the deformation decoder to predict deformation parameters $T=\{(\boldsymbol{R}_i,\boldsymbol{t}_i)\}_{i=1}^{K_s}$ for $\mathcal{G}_s$.

\subsection{Bounded Maximum Mean Discrepancy} 
Though the shared canonical shape $\mathcal{C}$ solves the problem of correlating extrinsic features, it is difficult for EI-AE to figure out how to form such a canonical shape without any self-supervision, which would result in severe overfitting. The reason for this is the domain gap between input features $\mathbf{X}_t\sim\mathcal{X}$ and the synthesized features $\mathbf{Y}_t\sim\mathcal{Y}$, where $\mathcal{X}$ and $\mathcal{Y}$ are the domains of input and synthesized features, respectively. Inspired by domain adaptation techniques~\cite{MMD, What_Transferred_Dong_CVPR2020, CSCL_Dong_ECCV2020, Semantic_Transferable_Dong_ICCV2019}, we propose to bridge the domain gap between $\mathcal{X}$ and $\mathcal{Y}$ by minimizing a Maximum Mean Discrepancy (MMD)~\cite{gretton2012kernel} loss: 
\begin{equation}\label{equ:MMD}
\begin{split}
M(\mathbf{X}_t, \mathbf{Y}_t)= & \mathbb{E}_{\mathbf{x}_i, \mathbf{x}_j\in\mathbf{X}}[\kappa(\mathbf{x}_i, \mathbf{x}_j)] + \mathbb{E}_{\mathbf{y}_i, \mathbf{y}_j\in\mathbf{Y}}[\kappa(\mathbf{y}_i, \mathbf{y}_j)]\\  &-2\mathbb{E}_{\mathbf{x}_i\in\mathbf{X}, \mathbf{y}_j\in\mathbf{Y}}[\kappa(\mathbf{x}_i, \mathbf{y}_j)]  ,
\end{split}
\end{equation}
where we omit the superscripts $t$, and $\kappa$ is the Radial Basis Function (RBF) kernel with $\kappa_\sigma(x, y)=\exp(-\frac{1}{2\sigma^2}||x-y||^2).$ We follow previous methods~\cite{li2017mmd} to apply a linear combination of kernels with five scales $\kappa(x, y)=\sum_{i}^{5}\kappa_{\sigma_i}(x, y)$, where $\sigma_i\in\{1, \sqrt{2}, 2, 2\sqrt{2},4\}$.

The motivation of this loss is to encourage the feature distributions inside original feature domain $\mathcal{X}$ and synthesized feature domain $\mathcal{Y}$ to be more compact (\emph{i.e.}, the first two terms of Eq.~\eqref{equ:MMD}), and meanwhile drives the synthesized feature domain $\mathcal{Y}$ to align with the original feature domain $\mathcal{X}$ by minimizing the distribution divergence across these two domains (\emph{i.e.},  the last term of Eq.~\eqref{equ:MMD}). It has been proven that $M(\mathcal{X}, \mathcal{Y})\geq 0$ and the equality holds if and only if $\mathcal{X}=\mathcal{Y}$~\cite{gretton2012kernel}. However, with a limit number of samples from $\mathcal{X}$ and $\mathcal{Y}$, optimizing MMD to be 0 may corrupt the deformation information in $\mathbf{X}_t$ and $\mathbf{Y}_t$. Thus, inspired by hinge loss, we introduce a bounded maximum mean discrepancy loss, with the bound $\beta=0.01$ in all experiments:
\begin{equation}
\mathcal{L}_{\rm Feat}=\max(0, M(\mathbf{X}_t, \mathbf{Y}_t)-\beta).
\end{equation}

\subsection{Trace and propagation}

\begin{figure}[htbp]
	\centering
	\includegraphics[width=0.48\textwidth]{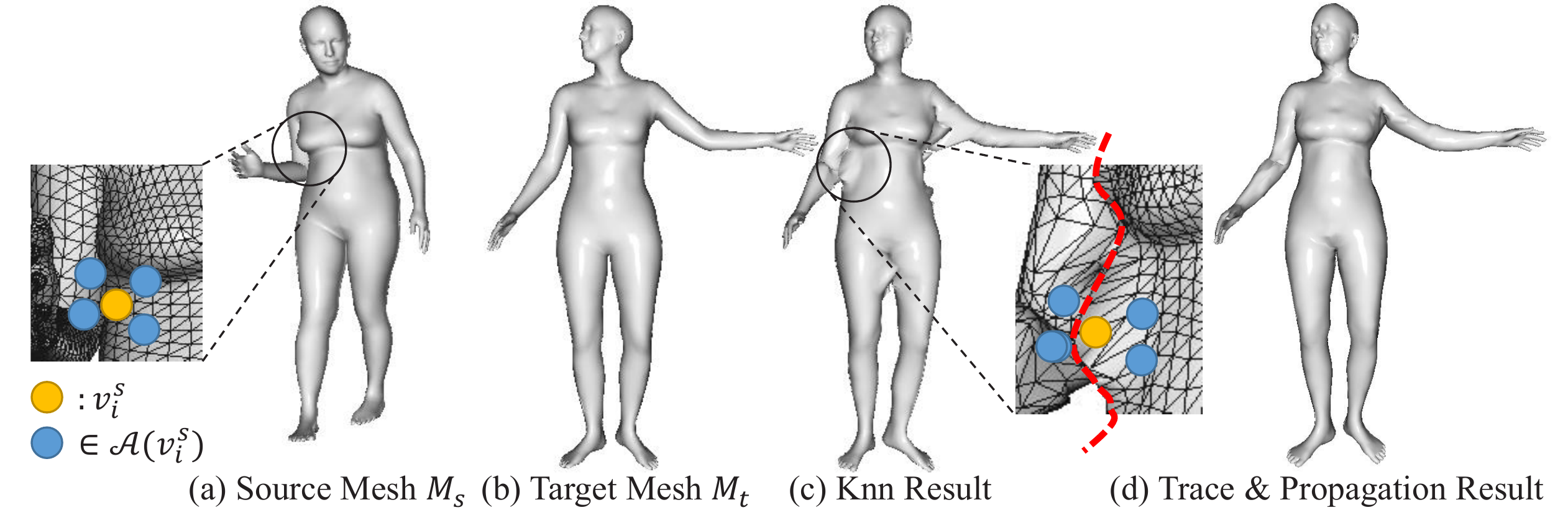}
	\caption{Comparison of using \emph{knn} in Euclidean space and our trace and propagation for node-to-vertex assignment.}\label{fig:knn}
\end{figure}

In Embedded Deformation (ED), a vertex $v^s_{i}$ on the source mesh $M_s$ is controlled by its $k$ nearest nodes $\mathcal{A}(v^s_{i})=\{\boldsymbol{g}_j^s\}_{j=1}^k$ in its deformation graph $\mathcal{G}_s$. Different from former methods with a fixed template~\cite{demea}, our method seeks to deform from source meshes with different topologies. Thus, the control nodes $\mathcal{A}(v^s_{i})$ for each vertex $v^s_{i}$ need to be redefined for each source mesh. Previous methods find $\mathcal{A}(v^s_{i})$ by searching $k$ nearest neighbour (\emph{knn}) of $v^s_{i}$ based on either geodesic or Euclidean distance. However, the geodesic distance is too heavy to compute, and the Euclidean distance may find wrong neighbouring nodes. For example, as shown in Fig.~\ref{fig:knn}, in $M_s$, the right arm is too close to the rib. Control nodes in arm are wrongly assigned to vertices in rib based on Euclidean distance, thus bring the ribs up with the lift of the arm. To this end, we propose a trace and propagation algorithm to find the control nodes $\mathcal{A}(v^s_{i})$, which is very efficient without any \emph{knn} search but can also improve the accuracy of the neighbours.

\begin{figure}[htbp]
	\vspace{-2pt}
	\centering
	\includegraphics[width=0.48\textwidth]{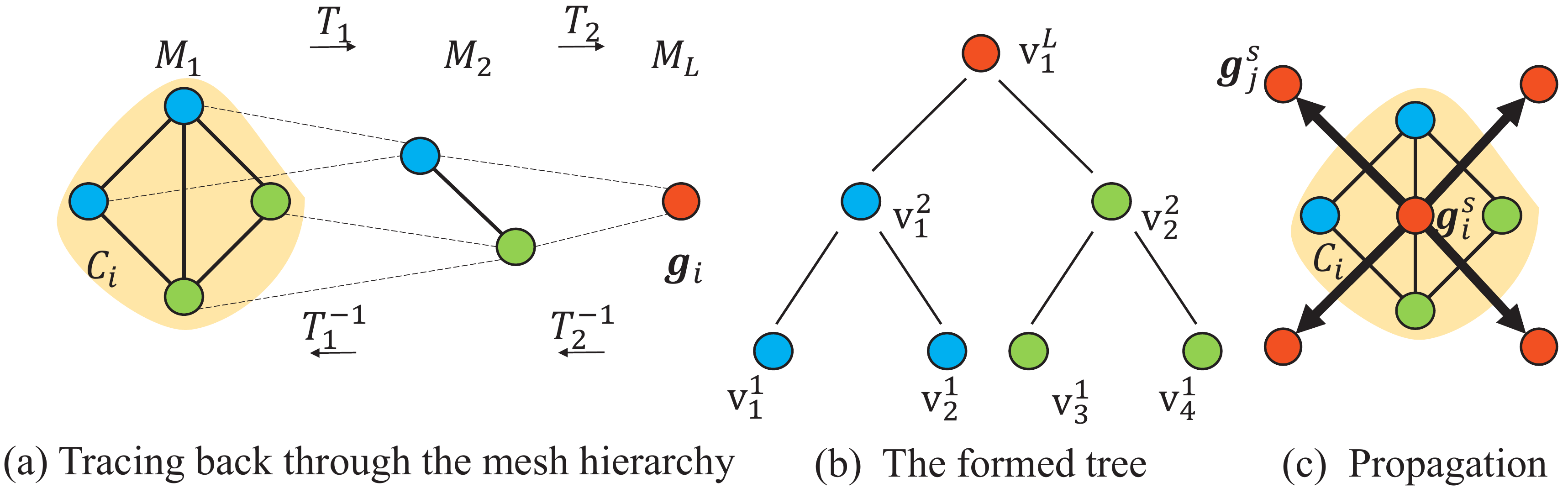}
	\caption{Example of the proposed \emph{trace and propagation} algorithm on a 3-level mesh hierarchy.}\label{fig:trace}
	\vspace{-2pt}
\end{figure}

Our main idea is to first trace back through the mesh hierarchy that is pre-constructed in the mesh encoder, and then propagate on the deformation graph to define more accurate neighbours. The mesh hierarchy $(M_1, M_2, \cdots, M_L)$, starting with the input mesh $M_1$ and ending with the coarsest deformation graph $M_L$, is generated by a series of poolings $(\mathcal{T}_1, \mathcal{T}_2, \cdots, \mathcal{T}_{L-1})$.
As shown in Figure~\ref{fig:trace}(a), a pooling operation $\mathcal{T}_l: \mathcal{V}^l\rightarrow \mathcal{V}^{l+1}$ maps a small partition of neighbouring vertices $\mathcal{O}=\{\mathbf{v}^l_i\}\subset\mathcal{V}^l,\mathbf{v}^l_i\in\mathbb{R}^3$ of mesh $M_l$ to one vertex $\mathbf{v}_j^{l+1}$ in $M_{l+1}$ by
\begin{equation}\label{equ:center}
\mathbf{v}_j^{l+1}=\sum_{\mathbf{v}^l_i\in\mathcal{O}} w^l_i\mathbf{v}^l_i,
\end{equation}
with the weight $w^l_i\geq0$ and $\sum w^l_i=1$. We further define a tracing operation $\mathcal{T}^{-1}_{l}$ as $\mathcal{T}^{-1}_l(\mathbf{v}_j^{l+1})=\mathcal{O}$, which restores the vertices in $\mathcal{O}$ from $\mathbf{v}_j^{l+1}$. 
Then, a node $\boldsymbol{g}^s_i$ in $M_L$ can trace back to $M_1$ layer by layer (\emph{i.e.}, $C_i=\mathcal{T}_1^{-1}\circ\mathcal{T}_2^{-1}\circ...\circ\mathcal{T}_{L-1}^{-1}(\boldsymbol{g}^s_i)$) to form a tree, as shown in Figure~\ref{fig:trace}(b), where $C_i$ is the leave set of the tree and contains all the original vertices in $M_1$ that are merged into $\boldsymbol{g}^s_i$. As shown in Figure~\ref{fig:trace}(c), $C_i$ can be regarded as a cluster of neighbouring vertices. According to Eq.~\eqref{equ:center}, since $\boldsymbol{g}^s_i$ is derived by a series of interpolation of vertices in $C_i$, $\boldsymbol{g}^s_i$ should lie near the center of $C_i$. For instance, if all $w_i$ are equal then $\boldsymbol{g}^s_i$ is the center of $C_i$. Thus, we regard $\boldsymbol{g}^s_i$ as the first control node for vertices in $C_i$, and define the rest of the control nodes for vertices in $C_i$ as the neighbours of $\boldsymbol{g}^s_i$ in deformation graph $\mathcal{G}_s$ (\emph{i.e.}, $\mathcal{A}(v^s_{i})=\{\boldsymbol{g}^s_i\}\cup\{ \boldsymbol{g}^s_j|(i,j)\in\mathcal{E}_s \}$).

The trace operation can be easily implemented as a look-up operation with computation complexity $O(N)$, where $N$ is the number of vertices in $M_{1}$. As shown in Figure~\ref{fig:knn}(d), with our proposed trace and propagation algorithm, the deformed mesh is smoother without any artifacts.

\subsection{Formulation}\label{loss}

We train our proposed UD$^2$E-Net without any supervision of ground-truth correspondences. Formally, the formulation objective consists of three main terms:
\begin{equation}
\mathcal{L}=\mathcal{L}_{\rm Align}+\mathcal{L}_{\rm Smooth}+\lambda_f\mathcal{L}_{\rm Feat},
\end{equation}
where $\mathcal{L}_{\rm Align}$ drives the source shape to align with the target, $\mathcal{L}_{\rm Smooth}$ regularizes the network to learn reasonable deformation based on smoothness and local rigidity. The third term is designed for EI-AE to align the feature space. The balanced weight $\lambda_f$ is set as 0.008 for all experiments.

For $\mathcal{L}_{\rm Align}$, we minimize the symmetric \emph{Chamfer Distance} $\mathcal{L}_{\rm Ch}$ between the vertices set $V_s'$, $V_t$ of deformed source $M_s'$ and the target mesh $M_t$.
However, for large deformation of articulated human bodies, $\mathcal{L}_{\rm Ch}$ may provide incorrect correspondences and lead to a bad local minima. Inspired by \cite{cycle-consistent}, we define a cycle-consistent loss $\mathcal{L}_{\rm Cyc}$ to enforce cycle-consistent correspondences between $M_s$ and $M_t$, which indicates that a cycle of deformation $M_s\rightarrow M_t\rightarrow M_s$ should map a point $p\in V_s$ back to itself.
Note that, to calculate this loss, \cite{cycle-consistent} forwards $M_s$ and $M_t$ twice for the deformations $M_s\rightarrow M_t$ and $M_t\rightarrow M_s$, which could be a large computational cost. While our Siamese pair of mesh encoders only forward the inputs once since they share the same parameters. Therefore, the objective $\mathcal{L}_{\rm Align}$ is:
\begin{equation}
\mathcal{L}_{\rm Align}=\mathcal{L}_{\rm Cyc}+\mathcal{L}_{\rm Ch}.
\end{equation}

To regularize the deformation, motivated by \cite{3d-coded}, we introduce an edge length term $\mathcal{L}_{\rm Edge}$ to penalize edge length changes and a Laplacian term $\mathcal{L}_{\rm Lap}$ to preserve local geometric details. With the as-rigid-as-possible term $\mathcal{L}_{\rm ARAP}$ in Eq.~\eqref{arap} for surface smoothness, $\mathcal{L}_{\rm Smooth}$ is formulated as:
\begin{equation}
\mathcal{L}_{\rm Smooth}=\lambda_{\rm Edge}\mathcal{L}_{\rm Edge}+\lambda_{\rm Lap}\mathcal{L}_{\rm Lap}+\lambda_{\rm ARAP}\mathcal{L}_{\rm ARAP},
\end{equation}
where the balanced weights $\lambda_{\rm Edge},\lambda_{\rm Lap},\lambda_{\rm ARAP}$ are set as $0.005$.

\section{Experiments}
\subsection{Dataset}
To validate the effectiveness of our model in both shape correspondence and shape reconstruction tasks, we evaluate the proposed UD$^2$E-Net on several challenging datasets, \ie, watertight meshes (DFAUST~\cite{dfaust}, SURREAL~\cite{surreal}, CoMA~\cite{coma}) and real scans from FAUST~\cite{FAUST}.

\textbf{FAUST}~\cite{FAUST} consists of 100 watertight meshes with ground-truth correspondences for training and 200 meshes for testing. The test data are real scans with noise and holes. \textbf{DFAUST}~\cite{dfaust} consists of sequences of meshes. There are 41220 watertight meshes in total. We follow the train-test split in~\cite{demea} for fair comparison. \textbf{SURREAL}~\cite{surreal} is a synthetic dataset based on SMPL~\cite{smpl} model. We follow~\cite{3d-coded} to generate 230k meshes including 200k human poses from~\cite{surreal} and 30k bent poses that can be unrealistic. We also sample 23k samples with 20k from~\cite{surreal} and 3k bent shapes to form a smaller training set. The testing set contains 200 meshes sampled from~\cite{surreal}. \textbf{CoMA}~\cite{coma} contains $17794$ watertight meshes of human faces spanning 12 subjects in 12 different expressions. The train-test split is also defined as in~\cite{demea}.

For all human body datasets, the meshes share the same topology with 6890 vertices. We decimate these meshes into 2757 vertices using Quadratic Error Metrics~\cite{qem} for training. When testing, the network also takes paired decimated meshes as input but output a high-resolution mesh by directly applying the predicted deformation parameters on the high-resolution source mesh. By doing this, we try to highlight that UD$^2$E-Net is capable of decoupling the number of deformation parameters from mesh resolution. Moreover, this also greatly accelerates the training process.

\subsection{Implementation Details}

\noindent
\textbf{Siamese Mesh Encoder.}
Our Siamese mesh encoder consists of graph convolution layers and graph pooling layers defined on a mesh hierarchy, which is constructed with Graclus~\cite{graclus} algorithm. We apply three downsampling layers to explore four mesh levels. The vertex number of each mesh level is approximately half of the preceding level. In each mesh level, we apply two \emph{EdgeConvs}~\cite{dgcnn} with a residual connection for feature extraction.\\
\textbf{Deformation Decoder.}
The architecture of deformation decoder $D_d$ follows mesh encoder but with only one mesh level to further build connections between neighbour nodes $\boldsymbol{g}^s_i$. It encourages the predicted transformations $\boldsymbol{T}_i$ to be closely similar for smoothness. We use two multi-layer perceptrons that end with linear layers to regress rotation $\boldsymbol{R}_i$ and translation $\boldsymbol{t}_i\in \mathbb{R}^3$, respectively. For rotation, we employ a 6D over-parameterized representation~\cite{6Drot}, given by the first two rows of the rotational matrix.\\
\textbf{Training Strategy.}
Our UD$^2$E-Net is trained end-to-end from scratch. We train it via the formulation introduced in Sec.~\ref{loss} and utilize the Adam optimizer with a fixed learning rate of 0.001. When the cycle-consistent loss stops to decrease after a few epochs, we train UD$^2$E-Net via $\mathcal{L}_{\rm Align}=\mathcal{L}_{\rm Ch}$ for the rest of epochs. Specifically, we first train UD$^2$E-Net on SURREAL dataset (230k) for 10 epochs, followed by 20 epochs without $\mathcal{L}_{\rm Cyc}$. On SURREAL dataset (23k), CoMA and DFAUST, UD$^2$E-Net is trained with $\mathcal{L}_{\rm Cyc}$ for 20 epochs followed by 180 epochs without it. For the evaluation on FAUST scans, we follow~\cite{3d-coded, deprelle2019learning} to train UD$^2$E-Net on SURREAL dataset (230k) without finetuning on the FAUST training set. 

\subsection{Baseline and Evaluation Criterion}\label{para:baseline}
\noindent
\textbf{Baseline.} To demonstrate the advantages of the fine-grained fusion enabled by EI-AE over the common global feature, we build a strong baseline, \emph{global embedding} (GE), which directly concatenates source local features $\mathbf{X}_s$ with the target global feature $\mathbf{h}_t$ as deformation embeddings. See the supplemental for detailed architecture.

\noindent
\textbf{Evaluation Criterion.} 
We evaluate our method on shape reconstruction and shape correspondence tasks. The reconstruction error measures the distance between the input target shape and reconstructed shape by computing Chamfer distance ($\rm R_{Ch}$) or mean Euclidean distance between ground-truth point pairs ($\rm R_{Gt}$). Correspondence error compares mean Euclidean distance between deformed points ($\rm C_{De}$) or their projections on target ($\rm C_{Pr}$) and the ground-truth correspondences on target. For template-based methods, correspondences are built following \cite{3d-coded}.

\subsection{Comparison Results} 

\noindent
\textbf{Experiments on Watertight Meshes.} As shown in Table~\ref{table:SURREAL}, we first compare our method with state-of-the-art methods on synthetic SURREAL dataset. The comparing methods apply a template-based~\cite{3d-coded, deprelle2019learning, demea} or an autoencoder framework~\cite{3dmm, coma}. For shape reconstruction, we also employ a fixed template as the source mesh $M_s$ for testing. Different from~\cite{3d-coded, deprelle2019learning} that apply the same template shape in both training and testing, UD$^2$E-Net is trained to reconstruct from random source shapes. Thus the comparison is not entirely fair. However, on SURREAL dataset (230k), our UD$^2$E-Net still outperforms all supervised methods 3D-CODED~\cite{3d-coded} and Elementary~\cite{deprelle2019learning}. With less sufficient training data (SURREAL 23k), the performance of 3D-CODED~\cite{3d-coded} and Elementary~\cite{deprelle2019learning} drops significantly up to $0.36$cm. In contrast, neither UD$^2$E-Net nor UD$^2$E-Net (GE) experience notable accuracy drop ($\leq0.11$cm) and they outperform all supervised methods, which proves the effectiveness of ED to alleviate strong dependence on data volume. For shape correspondence task, our UD$^2$E-Net outperforms the methods with a fixed template~\cite{3d-coded,deprelle2019learning} by a large margin ($0.25$$\sim$$0.72$cm for 230k and $1.19$$\sim$$1.32$cm for 23k), which shows the natural advantage of our template-free framework on shape correspondence task.

\begin{figure}[htbp]
	\centering
	\includegraphics[width=0.48\textwidth]{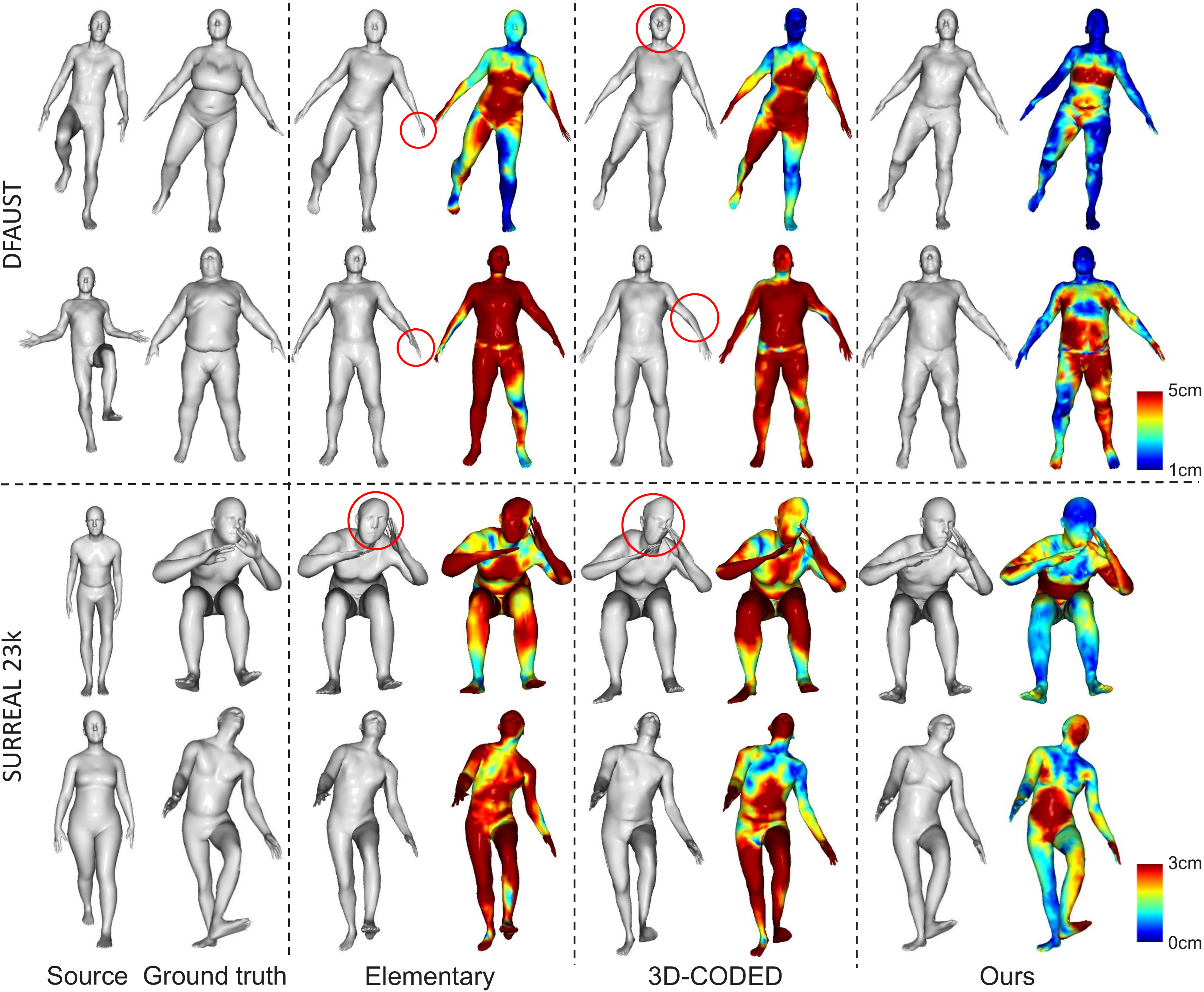}
	\caption{Qualitative comparison on DFAUST~\cite{dfaust} and SURREAL 23k with 3D-CODED~\cite{3d-coded} and Elementary~\cite{deprelle2019learning}. For each method, the left is the deformed source and right is its error map.}\label{fig:compare}
	\vspace{-10pt}
\end{figure}

On DFAUST dataset, variation of human poses and shapes are relatively limited, which results in poor performance for deformation-by-translation methods~\cite{3d-coded, deprelle2019learning} with $\rm{C_{De}}\geq6.46$cm and $\rm{R_{Gt}}\geq4.77$cm, since they fail to form a latent deformation space. On the contrary, with ED, UD$^2$E-Net holds high accuracy of 2.41cm and 2.17cm for $\rm{C_{De}}$ and $\rm{R_{Gt}}$. When pretrained on 230k samples from SURREAL, UD$^2$E-Net achieves $1.84$cm for reconstruction and outperforms all competing methods even the autoencoder N. 3DMM~\cite{3dmm} ($1.99$cm). Our method also suits for finer deformation of faces in CoMA dataset. As shown in Table~\ref{tab:coma}, the proposed UD$^2$E-Net (0.78mm) also outperforms the supervised DEMEA~\cite{demea} (0.81mm).
Moreover, we also show qualitative comparison with 3D-CODED~\cite{3d-coded} and Elementary~\cite{deprelle2019learning} in Figure~\ref{fig:compare}. The comparison methods fail to capture local geometric details with a global feature, thus results in local distortion, such as in hands, arms and especially faces where rich geometric details exist.
\begin{table}[htbp]

	\begin{center}
		\setlength{\tabcolsep}{0.5mm}
		\begin{tabularx}{8cm}{c|c|XXXX}
			\hline
			Data	&	Methods&$\rm{R_{Gt}}$& $\rm {R_{Ch}}$	&$\rm {C_{De}}$	& $\rm {C_{Pr}}$\\	\hline
			\multirow{8}*{230k}  &	3D-CODED~\cite{3d-coded} 	
			&1.80			&		1.03		&	2.63	&		2.30			\\
			& Elementary~\cite{deprelle2019learning}		
			&\textbf{1.67}			&		0.89		&	2.19	&	1.87			\\
			& Unsup. 3D-CODED~\cite{3d-coded} 	
			&	9.43			&	1.76	&		9.30		&	9.27		\\
			& Ours-GE 			&	1.96	&		1.00 		&	2.23	&	1.88				\\
			& Ours-w/oTP 			&	3.25	&		1.15			&	2.85		&	3.00	\\
			& Ours-w/oMMD 		& 	2.11	&		1.06 		&	2.47	&	2.12	\\
			& Ours-3D  			& 	1.71		&	\textbf{0.86}		&  2.02&1.96\\
			& Ours 				&	\textbf{1.67}			&	0.87				&	\textbf{1.91}		&		\textbf{1.62}
			\\\hline
			\multirow{8}*{23k} &	3D-CODED~\cite{3d-coded}	
			&	2.10		&		1.18			&	3.36			&	3.01	\\
			&	Elementary~\cite{deprelle2019learning}	&	2.03		&	1.15				&	3.27				&	2.91\\
			& Unsup. 3D-CODED~\cite{3d-coded} 	
			&	9.32	&	1.86	&		9.59&9.43	\\
			& Ours-GE 		&	2.22	&	1.07	&	2.61	&2.28	\\
			& Ours-w/oTP 		&	2.18	&	1.04		&	3.08		&	2.80	\\
			& Ours-w/oMMD 		& 	2.16	&	1.05	&	2.47	&2.13	\\
			& Ours-3D  		& 	1.83	&	0.93	&	2.15	&1.84	\\
			& Ours 				&	\textbf{1.78}		&		\textbf{0.911}		&	\textbf{2.04}		&	\textbf{1.72}		\\
			\hline
		\end{tabularx}
	\end{center}
	\caption{Performance comparison on synthetic SURREAL dataset~\cite{surreal}. `230k' and `23k' denote the amount of training data.}
	\label{table:SURREAL}
	\vspace{-5pt}
\end{table}
\begin{table}[htbp]
	\begin{center}
		\setlength{\tabcolsep}{2.5mm}
		\label{dfaust}
		\begin{tabularx}{8cm}{l|c|cc}
			\hline
			&	Methods			& 	$\rm{C_{De}}$ (cm)	& $\rm{R_{Gt}}$ (cm)			\\	\hline
			\multirow{7}*{Original}
			& N. 3DMM~\cite{3dmm} 	&	-		&		1.99					\\
			& CoMA~\cite{coma} 	&	-		&		8.4					\\
			& DEMEA~\cite{demea} 		&	-		&	2.23						\\
			&3D-CODED~\cite{3d-coded} 	&	6.46		&	4.81		\\
			&Elementary~\cite{deprelle2019learning} 	&	6.51		&	4.77		\\
			&Ours-GE 	&	3.03		&	2.57							\\	
			&Ours  	&	2.41		&		2.17				\\ \hline
			\multirow{4}*{230k} 
			&3D-CODED~\cite{3d-coded}		&	4.91	& 3.55	\\
			&Elementary~\cite{deprelle2019learning} 	&	4.58		&	3.36		\\
			&Ours-GE 	&	2.24		&			1.95					\\	
			& Ours		&	\textbf{2.07}		&			\textbf{1.84}				\\
			
			\hline
		\end{tabularx}
	\end{center}
	\caption{Performance comparison on DFAUST dataset. `Original' denotes only training on DFAUST\cite{dfaust}. `230k' means finetuning on DFAUST with pretrained network on SURREAL (230k).}
	\vspace{-5pt}
\end{table}

\begin{table}[htbp]
	\begin{center}
		\setlength{\tabcolsep}{3.mm}
		\begin{tabular}{c|c|c|c}
			\hline
			&	DEMEA 		&	N. 3DMM	    &	Ours 				\\ \hline
			$\rm R_{Gt}$ (mm) 	&	 	0.81	&    \textbf{0.71}		&	 0.78				\\ \hline
		\end{tabular}
	\end{center}
	\caption{Performance comparison on CoMA dataset~\cite{coma}.}~\label{tab:coma}
	\vspace{-25pt}
\end{table}

\noindent\textbf{Experiments on Real Scans.}
We then test our UD$^2$E-Net on more challenging scan dataset, where missing parts (\eg, feet, hands, etc.) may exist due to self-occlusion. These meshes have approximately 170000 vertices. Instead of directly sending them into the network, we follow the pre-processing strategy for training to first downsample the scanned meshes to 2757 vertices. Since the testing meshes may have topological changes caused by self-contact, UD$^2$E-Net may fail to deform parts with topological noise. Thus, given a pair of testing meshes, we also utilize a watertight mesh as a template to build correspondences as in~\cite{3d-coded,deprelle2019learning}. Though a template is employed, our method is still naturally superior to template-based methods since we can freely choose better templates. In practice, we choose the null pose of each test human as template. Similar to~\cite{3d-coded,deprelle2019learning}, we optimize global feature of the target to obtain the final reconstruction. Some comparison results are shown in Table~\ref{faust}. On the `Inter' challenge, our method achieves 3.086cm, which outperforms unsupervised methods significantly by 24$\%$$\sim$37$\%$ and is comparable to supervised methods. On the `Intra' challenge, our method achieves \textbf{1.512}cm, which even outperforms state-of-the-art supervised method~\cite{deprelle2019learning} by 13$\%$.

\begin{table}
	\begin{center}
		\setlength{\tabcolsep}{2.0mm}
		\begin{tabular}{l|c|c|c}
			\hline
			&	Methods			& 	Inter. 	& 	Intra. 			\\	\hline
			\multirow{3}*{Sup.}
			& FMNet~\cite{FMNet}		&	4.83		&	2.44\\
			&3D-CODED~\cite{3d-coded} 	&	2.878		&	1.985		\\
			& Elementary~\cite{deprelle2019learning}		&	\textbf{2.58}		&		\textbf{1.742}	\\					
			\hline
			\multirow{5}*{Unsup.}
			&Unsup. 3D-CODED~\cite{3d-coded} 	&	4.88		&	-		\\
			& LBS Autoencoder~\cite{li2019lbs}			&	4.08		&		2.161					\\
			& Halimi~\etal\cite{halimi2019unsupervised}		&	4.883		&	2.51\\
			& Ginzburg~\etal\cite{ginzburg2020cyclic}	&	4.068		&	2.12						\\
			& Ours		&	\textbf{3.086}		&				\textbf{1.512}			\\
			
			\hline
		\end{tabular}
	\end{center}
	\caption{Performance comparison on Faust dataset. `Sup.' and `Unsup.' indicate the method is supervised or unsupervised.}		\label{faust}
	\vspace{-15pt}
\end{table}

\noindent\textbf{Ablation Study.}
As shown in Table~\ref{table:SURREAL}, to demonstrate the effectiveness of each proposed component, we compare our final UD$^2$E-Net with several variants including: 1) the baseline (Ours-GE), 2) UD$^2$E-Net without MMD (Ours-w/oMMD), 3) UD$^2$E-Net with \emph{knn} instead of trace and propagation (Ours-w/oTP), and 4) UD$^2$E-Net 3D that applies 3D canonical shape in the EI-AE (Ours-3D).
Without the proposed EI-AE, the performance drops 13$\%$ in shape correspondence, which demonstrates the indispensable role of fine-grained feature fusion in deformation embedding. Moreover, the employed bounded MMD is critical to align the distributions between synthetic target features generated by EI-AE and the original features, and guide EI-AE to form the shared canonical shape. Without bounded MMD, the performance drops 23$\%$$\sim$17$\%$. Without accurate control nodes provided by the proposed trace and propagation algorithm, the performance drops 33$\%$$\sim$51$\%$ on shape correspondence. With 3D canonical shape $\mathcal{C}$, UD$^2$E-Net performs slightly worse than using a 10D shape, which shows that UD$^2$E-Net is not sensitive to shape dimension.\medskip

\noindent
\subsection{Effect of the Extrinsic-Intrinsic Autoencoder}
\begin{figure}[htbp]
	\centering
	\includegraphics[width=0.48\textwidth]{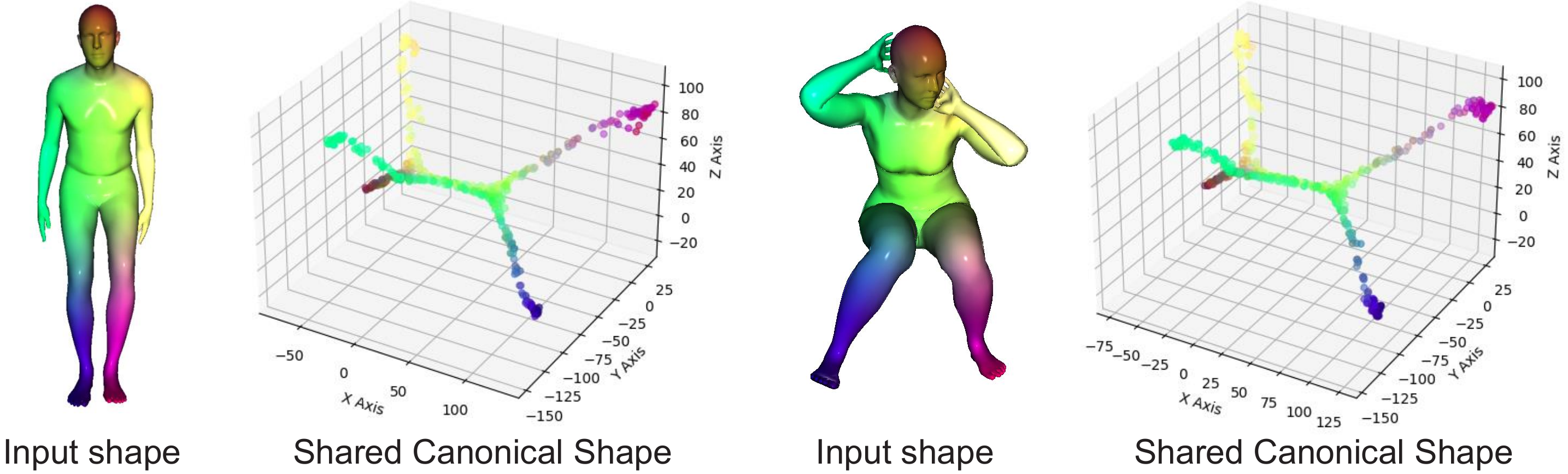}
	\caption{Visualization of the learned canonical shape in EI-AE. Same color indicates correspondence.}\label{fig:Study_IB:b}
	\vspace{-15pt}
\end{figure}

\begin{figure}[htbp]
	\centering
	\includegraphics[width=0.48\textwidth]{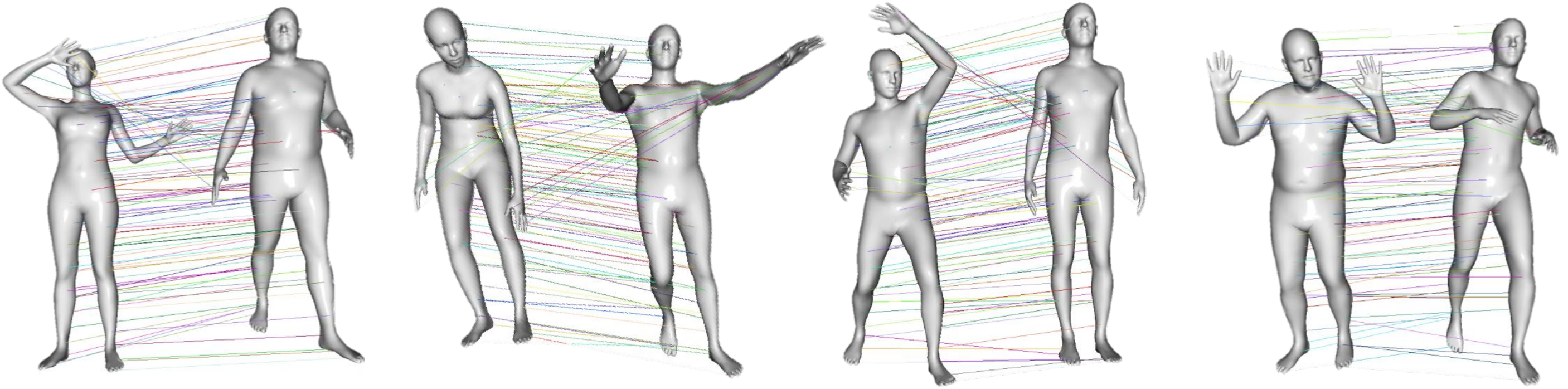}
	\caption{Results of point matches based on the coordinates $\mathbf{c}_i$.}\label{fig:Study_IB:a}
	\vspace{-10pt}
\end{figure}

We study the effect of EI-AE on dense deformation embedding by exploring how the intrinsic coordinates distributed inside the shared canonical shape and the effect of shared coordinates across different bodies. We first visualize the distribution of 3D coordinates in Figure~\ref{fig:Study_IB:b}, which shows that the learned canonical shape presents the shape of the skeleton of a body, whose limbs and head are consistent with the original human body. Though input bodies are highly different, the derived canonical shapes are nearly identical. Between these two shapes, we build correspondences by finding nearest neighbour based on coordinates $\mathbf{c}_i$. As shown in Figure~\ref{fig:Study_IB:a}, the proposed EI-AE is capable of mapping corresponding vertices across different bodies into consistent coordinates in $\mathcal{C}$, though without any supervision of correspondences. Above results have proved our assumption that the EI-AE disentangles intrinsic location from geometric information, which enables the synthesized $\mathbf{Y}_t$ to focus on capturing geometric information, thus boosting the deformation quality. Moreover, since the canonical shape implies correspondences, it can help with unsupervised deformation learning by enabling the network to learn a deformation space that complies with these correspondences.


\subsection{Applications}

This subsection validates that our UD$^2$E-Net could be successfully extended into several applications, \eg, shape retrieval and human pose transfer. More results including shape interpolation are included in the supplemental.

\noindent
\textbf{Template Retrieval.}
For shape reconstruction problem, choosing a more similar template can obviously improve the performance. Given a target shape from the test set, we further try to retrieve the source shape from training set such that the cosine distance between their global embeddings are minimized. As shown in Table~\ref{tab:retrieval}, where `Fixed' and `Retrieved' denote using a fixed or retrieved template, our model provides an additional improvement of 4.8$\%$ on DFAUST and a significant promotion of 37.1$\%$ on SURREAL. As shown in Figure~\ref{fig:retrieval}, the retrieved shapes have similar poses with the query shapes.
\begin{figure}
	\centering
	\includegraphics[width=0.48\textwidth]{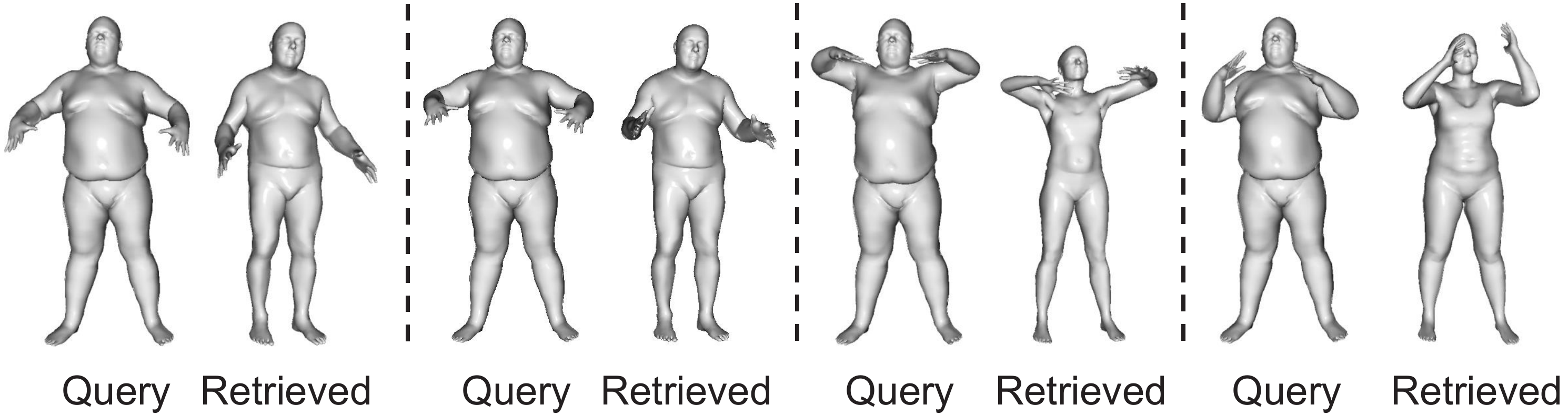}
	\caption{Results of shape retrieval on DFAUST dataset, where the query shapes are on the left and retrieved shapes are on the right.}\label{fig:retrieval}
	\vspace{-5pt}
\end{figure}
\begin{table}
	\begin{center}
		\setlength{\tabcolsep}{1.8mm}
		\begin{tabular}{c|c|c|c|c}
			\hline
			&	\multicolumn{2}{c|}{DFAUST}		&	\multicolumn{2}{c}{SURREAL (23K)}	\\ \cline{2-5}
			&	Ours (GE) 		&	Ours 		&  	Ours (GE)		&	Ours	    \\ \hline
			Fixed 		&	 	1.95		&	 1.84		& 		2.22		&    1.78		\\
			Retrieved 	&	 	1.87		&	 1.752		& 		1.67		&    1.12		\\ 
			$\Delta$	&		-4.1$\%$	&	-4.8$\%$	& 		-24.8$\%$	&	-37.1$\%$		\\\hline
		\end{tabular}
	\end{center}
	\caption{Performance improvement brought by template retrieval.}~\label{tab:retrieval}
	\vspace{-15pt}
\end{table}

\begin{figure}
	\vspace{-5pt}
	\centering
	\label{fig:subfig:a} 
	\includegraphics[width=0.48\textwidth]{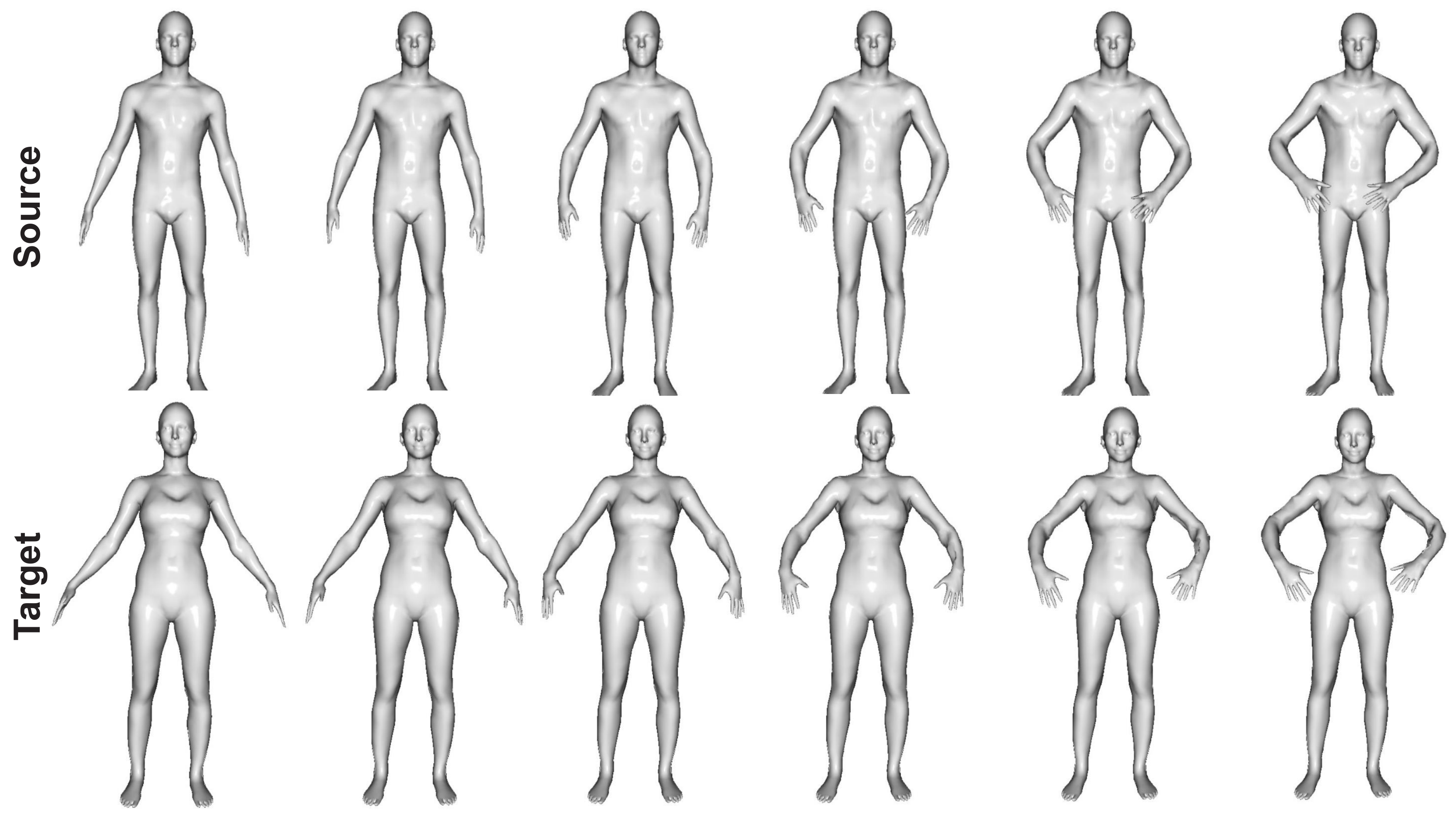}
	\caption{Deformation transfer results.}\label{fig:defomation_transfer} 
	\vspace{-15pt}
\end{figure}

\noindent
\textbf{Human pose transfer.}
The learned latent space also allows to transfer poses for human bodies. Given the meshes of human body $A$ in two poses $A_0, A_1$, we represent their embedded global features as $\mathbf{h}_0, \mathbf{h}_1$. Then given another human body $B_0$, which is of the same pose as $A_0$ with its embedding denoted as $\mathbf{h}_0'$, we seek a target shape $B_1$ performing same pose as $A_1$. The deformation is transferred by computing the embedding of $B_1$ as $\mathbf{h}_1'=\mathbf{h}_1-\mathbf{h}_0+\mathbf{h}_0'$. Then $B_1$ can be derived by decoding $\mathbf{h}_1'$. We evaluate the effectiveness of our model in a more challenging situation by transferring interpolated poses. As the qualitative results shown in Figure~\ref{fig:defomation_transfer}, the model trained on SURREAL dataset could be employed to transfer poses in FAUST dataset.

\section{Conclusion}
We have presented UD$^2$E-Net, an end-to-end network that allows learning the deformation between arbitrary shape pairs from dense local features in an unsupervised manner. With the Extrinsic-Intrinsic Autoencoder guided by bounded MMD, the network can explore both extrinsic property for fine-grained deformation reasoning and intrinsic property for matching by forming a shared canonical shape. Moreover, the trace and propagation algorithm can improve both efficiency and quality of the deformation graph. UD$^2$E-Net shows significant improvement over unsupervised methods and even outperforms supervised methods. We hope that our work inspires the field of neural deformation and can be extended to more challenging applications, \eg, non-rigid reconstruction and tracking.

{\small
	\bibliographystyle{ieee_fullname}
	\bibliography{main}
}
\appendixpageoff
\appendixtitleoff
\renewcommand{\appendixtocname}{Supplementary material}
\begin{appendices}

\title{\emph{Supplementary Material:} \\Dense Deformation Embedding Network for Template-Free \\ Shape Correspondence}
\emptythanks
\author{}
\date{}



\maketitle


In this supplementary material, we first provide detailed architecture of the Extrinsic-Intrisic Autoencoder and the global embedding baseline, in Sec.~\ref{architecture}. We then describe the details on construction of the mesh hierarchy in Sec~\ref{hierarchy}. In Sec.~\ref{EIAE}, we provides more analysis on the effect of the Extrinsic-Intrisic Autoencoder. Sec.~\ref{ablation} provides more ablation studies on employed losses. We evaluate the model complexity in Sec.~\ref{Timing}. Sec.~\ref{methods} provides more details on the methods that we compare with in the experiments of the main paper. In Sec.~\ref{qualitative}, we show more qualitative results on mesh deformation. In Sec.~\ref{application}, we provide more visualization results for applications of our model in human pose transfer, shape retrieval and shape interpolation.

\section{Network Architecture}\label{architecture}

\subsection{Extrinsic-Intrinsic Autoencoder}
\begin{figure}[htbp]
	\centering
	\includegraphics[width=0.48\textwidth]{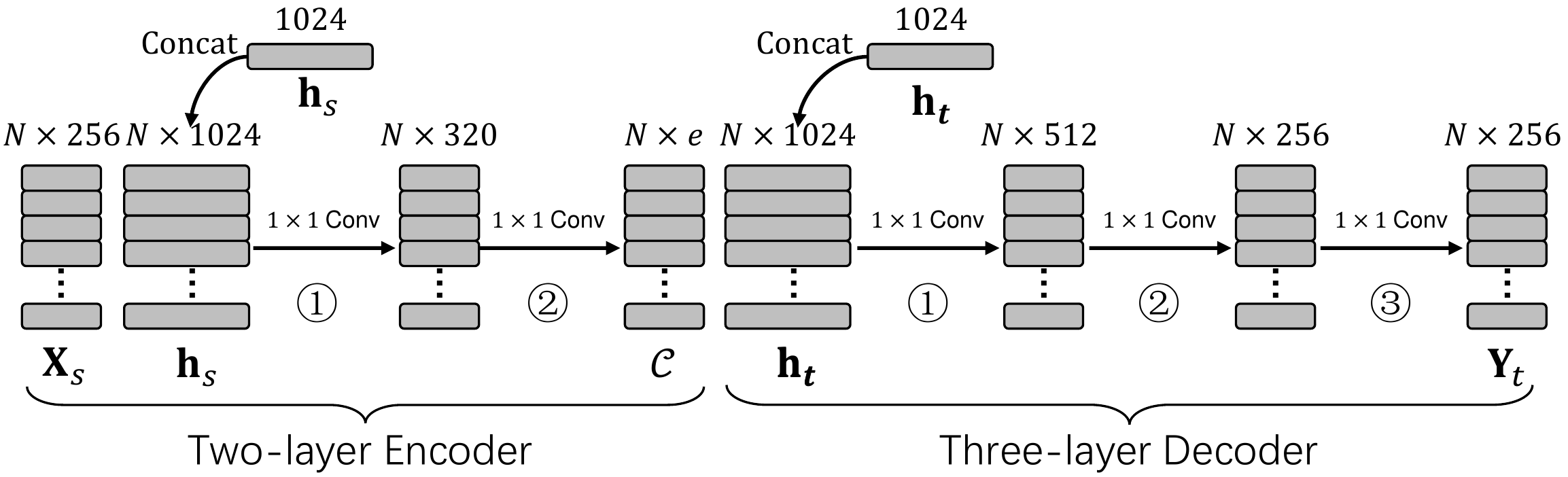}
	\caption{The architecture of the Extrinsic-Intrinsic Autoencoder.}\label{fig:EI-AE}
\end{figure}

As shown in Figure~\ref{fig:EI-AE}, we present the detailed architecture of our proposed Extrinsic-Intrinsic Autoencoder (EI-AE). It consists of a two-layer encoder and a three-layer decoder. We apply \emph{BatchNorm} and \emph{ReLU} after each convolution layer, except for the bottleneck layer where the canonical shape $\mathcal{C}$ is directly output by a convolution layer. Without specification, we apply a 10D shared canonical shape $\mathcal{C}$ (\ie, $e=10$). We also apply 3D shared canonical shape in ablation study.

\subsection{Baseline}
\begin{figure}[htbp]
	\centering
	\includegraphics[width=0.48\textwidth]{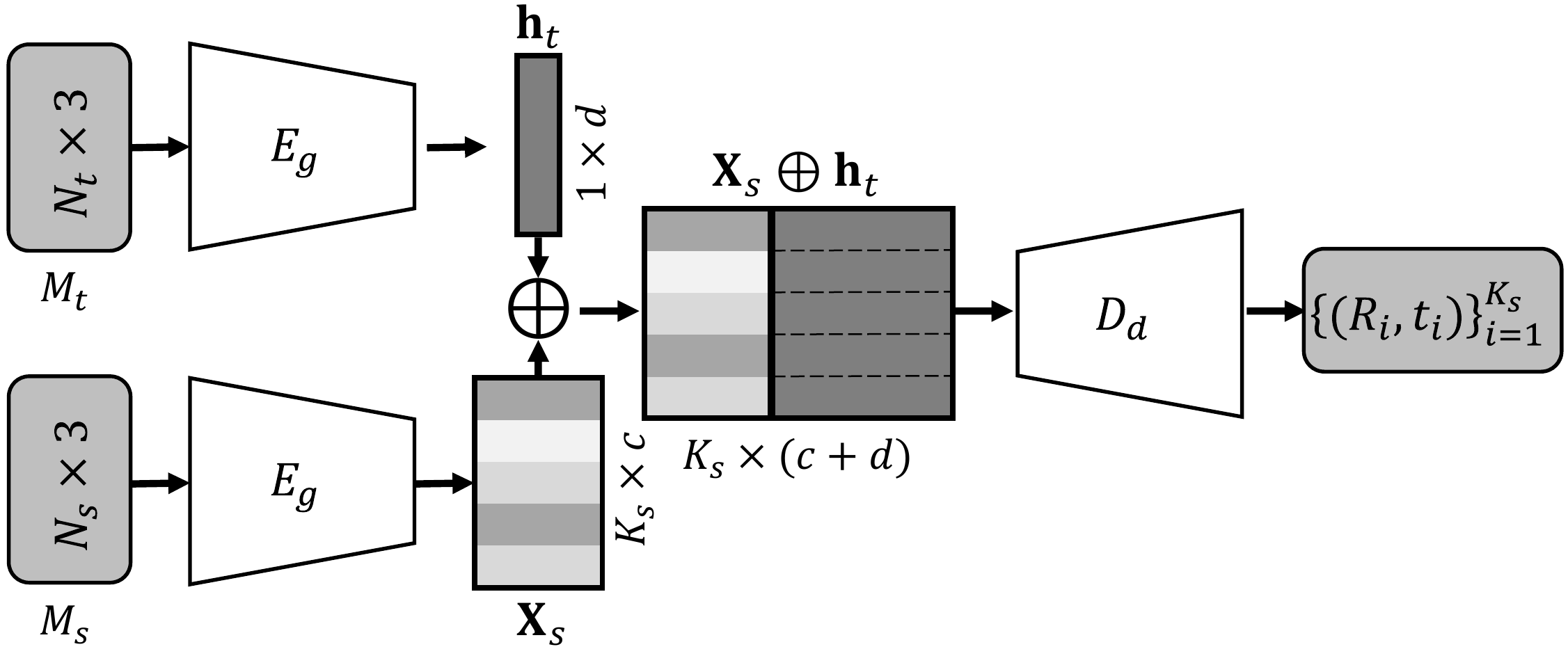}
	\caption{The architecture of the \emph{global embedding} baseline.}\label{fig:baseline}
\end{figure}

Figure~\ref{fig:baseline} illustrates the architecture of our \emph{global embedding} (GE) baseline. It employs the same Siamese mesh encoder $E_g$ and deformation decoder $D_d$ as our proposed UD$^2$E-Net. The only difference between them is that the GE baseline removes the EI-AE and directly concatenates source local features $\mathbf{X}_s$ with the target global feature $\mathbf{h}_t$ as deformation embeddings.

\section{Mesh Hierarchy}\label{hierarchy}

\begin{figure}[htbp]
	\centering
	\includegraphics[width=0.48\textwidth]{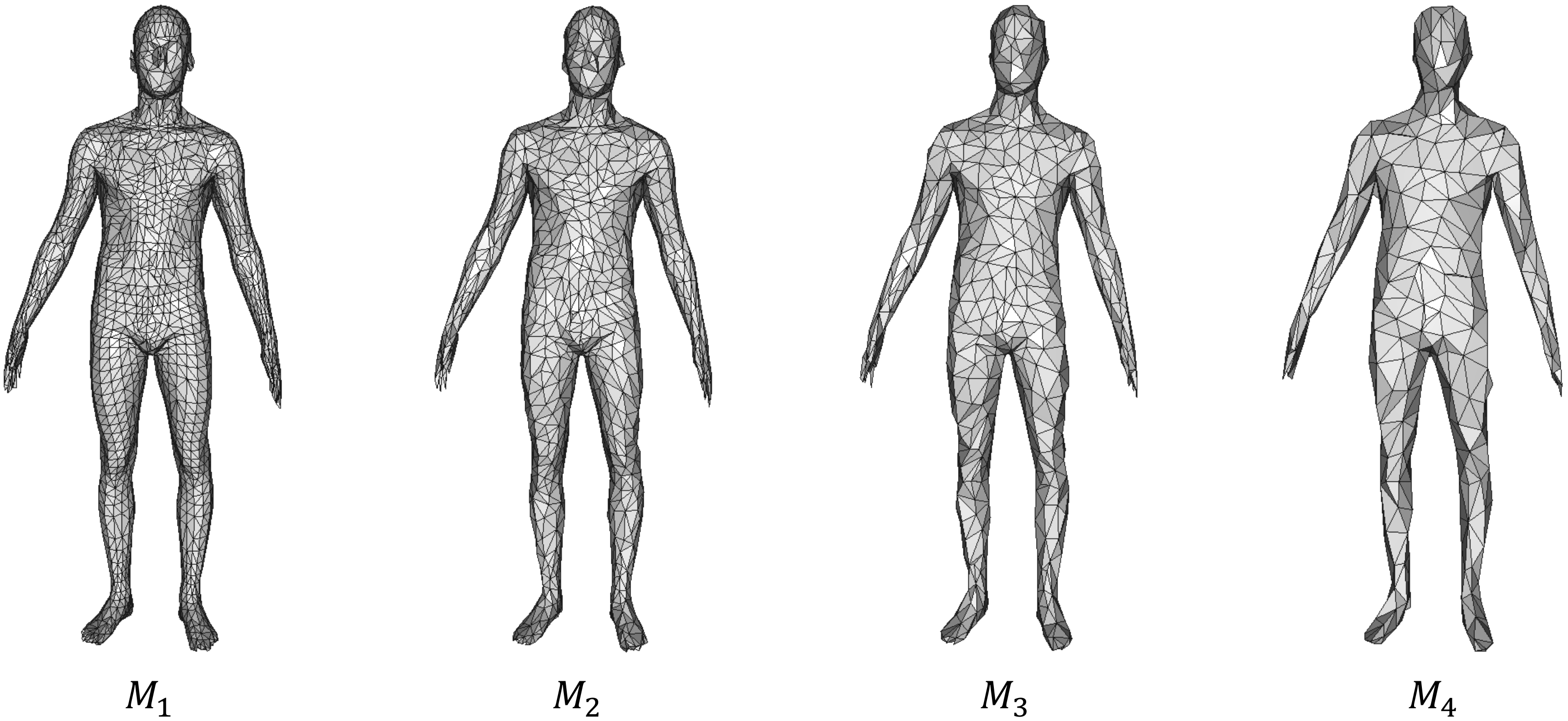}
	\caption{The mesh hierarchy generated by Graclus algorithm~\cite{graclus}.}\label{fig:meshhierarchy}
\end{figure}

As shown in Figure~\ref{fig:meshhierarchy}, we utilize the Graclus algorithm~\cite{graclus} to construct the mesh hierarchy ($M_1,M_2,M_3,M_4$), where the coarsest mesh $M_4$ is the deformation graph $\mathcal{G}$ and $M_1$ is the input mesh. We construct mesh hierarchy for two main purposes. First, the meshes are downsampled to enable graph pooling on feature maps, which can not only reduce number of parameters and computational complexity by scaling down the size of feature maps but also enable \emph{Graph Convolution Networks} (GCNs) to better aggregate information with enlarged receptive field sizes and learn hierarchical representations. Second, we downsample the input mesh to utilize the derived low-resolution mesh as the deformation graph $\mathcal{G}$.

Recently, DEMEA~\cite{demea} also defines graph convolutions on a mesh hierarchy to learn deformation embeddings for a deformation graph. However, in~\cite{demea}, the mesh hierarchy is computed once prior to the training process using QEM based methods. This requires all training data to share the same topology, which vastly limits its application. We address this challenge by using the Graclus algorithm to downsample the graph in real-time simultaneously with the forwarding of the network, due to its high efficiency. Graclus traverses all nodes in the graph and in each step greedily merges two unmarked nodes that maximize the local normalized cut $w_{ij}(d_{i}^{-1}+d_{j}^{-1})$, then mark them as visited. When all nodes are visited, the graph has approximately half of the nodes. The graph hierarchy can be obtained by repeating this process until required resolution. Due to randomness of the algorithm, the derived mesh hierarchy can be updated dynamically, which also improves the generalization ability for GCNs.

Moreover, the vertex number of each mesh level is also not fixed. The vertex number of the input mesh is fixed as 2757, whereas the rest mesh levels have around 1480, 800, 430 vertices, respectively.

\section{Effect of the Extrinsic-Intrinsic Autoencoder}\label{EIAE}

\begin{figure}[htbp]
	\centering
	\includegraphics[width=0.48\textwidth]{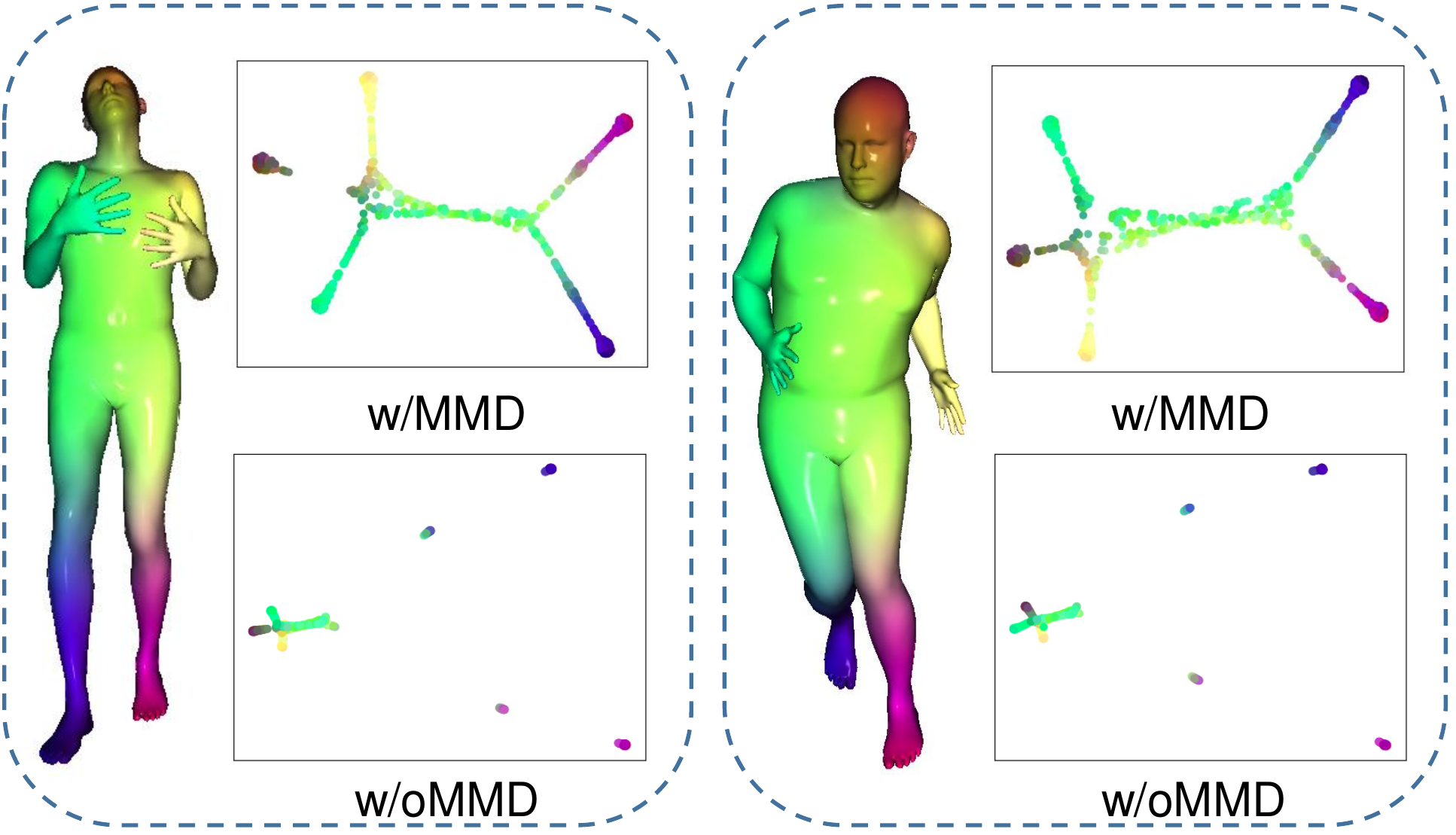}
	\caption{Visualization of 10D shared canonical shape $\mathcal{C}$ via t-SNE, where the left is the input shape and the top-right and bottom-right are the learned $\mathcal{C}$ with and without bounded MMD, respectively.}\label{fig:tsne}
\end{figure}

In this section, we provide more visualization results for the proposed EI-AE. As shown in Figure~\ref{fig:tsne}, we visualize the learned 10D shared canonical shape $\mathcal{C}$ via t-SNE. Similar to the 3D shared canonical shape, the 10D shared canonical shape also presents the shape of the skeleton of a body. However, without the guidance of the bounded \emph{Maximum Mean Discrepancy} (MMD), the EI-AE fails to form a compact canonical shape, which demonstrates the indispensable role of the introduced bounded MMD.

\section{More ablation studies}\label{ablation}
\subsection{Bounded Maximum Mean Discrepancy}
We conduct ablation studies about $\beta$ in Equation. 6 of the main paper among $\{0, 0.01, 0.1\}$ on SURREAL (230k) dataset. As shown in Table~\ref{tab:lambda}, the performance is better when $\beta>0$, and is not quite sensitive to the choice of $\beta$. It proves that the hinge loss prevents the feature corruption, as illustrated in Section 3.3 of the main paper.

\vspace{-8pt}
\begin{table}[htbp]
	\begin{center}
		\begin{tabularx}{5.2cm}{c|cccc}
			\hline
			$\beta$			&	0		&	0.1			&	0.01	\\ \hline
			$\rm{C_{DE}}$ (cm)	&	2.30	&	2.04		&	\textbf{1.91}	\\
			\hline
		\end{tabularx}
	\end{center}
	\vspace{-5pt}
	\caption{Ablation studies about $\beta$ on SURREAL (230k) dataset.}\label{tab:lambda}
	\vspace{-10pt}
\end{table}

\subsection{Cycle-Consistent Loss}
\begin{figure}[htbp]
	\centering
	\includegraphics[width=0.48\textwidth]{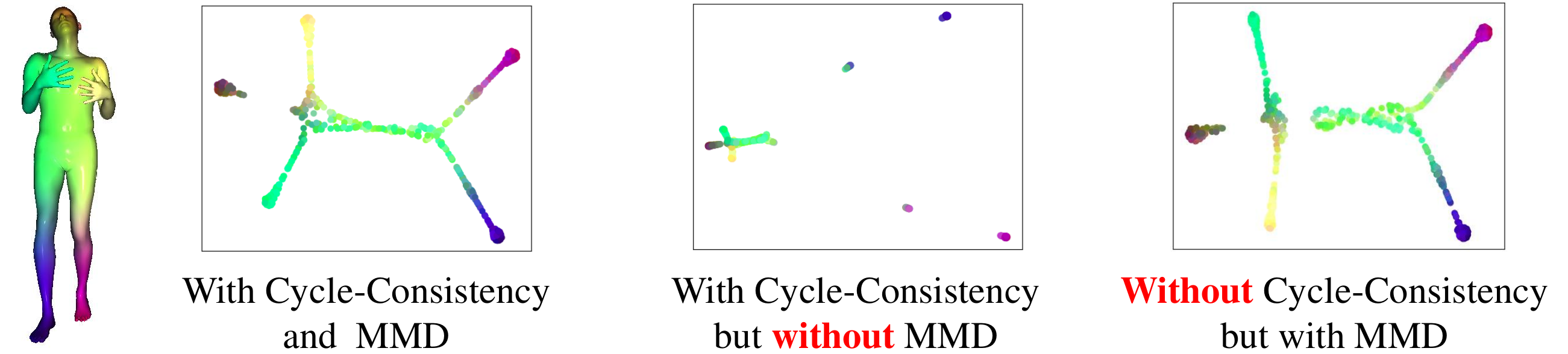}
	\caption{Representation visualization comparison with/without MMD and cycle-consistent loss.}\label{fig:rep}
\end{figure}

Both bounded MMD and the cycle-consistent loss can provide self-supervision for our UD$^2$E-Net. Here we discuss their effect on the construction of the canonical shape $\mathcal{C}$. We visualize the learned canonical shape in Figure~\ref{fig:rep}. Specifically, our model without MMD fails to form a compact canonical shape. However, discarding the cycle-consistent loss does not degenerate the representation, which proves the indispensable role of MMD over the cycle-consistent loss. 

\section{Timing and model complexity}\label{Timing}

We measure the run-time of UD$^2$E-Net on Faust Dataset. It takes 68.32ms to process a pair of watertight meshes with 6890 vertices. The measurements are performed on an NVIDIA TITAN XP GPU across 500 runs. The entire network has 3.26M parameters.

\section{Comparison Methods}\label{methods}

The methods we compare with can be divided into deformation-based methods~\cite{3d-coded,deprelle2019learning,li2019lbs,demea}, mesh autoencoders~\cite{coma, 3dmm} and spectral methods~\cite{FMNet, halimi2019unsupervised, ginzburg2020cyclic}. 

For deformation-based methods, \textbf{3D-CODED}~\cite{3d-coded} deforms a fixed template to align with the input target shape. It encodes the target shape globally into a 1024-D vector with a PointNet-like~\cite{pointnet} encoder, then predicts a deformed location for each point on the template by decoding the concatenation of the global vector and the point's location. The method is fully supervised with ground-truth correspondences. It also has an unsupervised variant, which we denote as Unsup$.$ 3D-CODED in Table.~1,4 of the main paper. \textbf{Elementary}~\cite{deprelle2019learning} extends 3D-CODED by automatically learning a better elementary structure from a shape collection for shape reconstruction and matching. It also deforms the input shape by predicting per-point locations and is fully supervised. In all experiments, we apply its 3D \emph{patch deformation} variant, since it achieves the best performance on Faust benchmark~\cite{FAUST}. DEMEA employs an embedded deformation layer based on~\cite{sumner2007embedded} to deform a fixed template to restore the input shape, which is also fully supervised. \textbf{LBS-Autoencoder}~\cite{li2019lbs} utilizes Linear Blending Skinning (LBS) for deformation, which is self-supervised with ground-truth joint rotation angles.

The mesh autoencoders~\cite{coma, 3dmm} encode the input mesh into a latent code with graph convolutions, and then directly decode it to restore the input mesh.

Spectral methods perform matching in the spectral domain. They are built upon a functional map representation to learn descriptors for matching. \textbf{FMNet}~\cite{FMNet} is supervised with ground-truth correspondences. \textbf{Halimi \etal}~\cite{halimi2019unsupervised} assume isometric deformations and remove the supervision by minimizing pair-wise geodesic distance distortions. \textbf{Ginzburg \etal}~\cite{ginzburg2020cyclic} introduce cyclic mapping, which can generalize to non-isometric deformation and achieves state-of-the-art performance among unsupervised methods.

Our proposed UD$^2$E-Net outperforms above supervised and unsupervised methods on SURREAL and DFAUST dataset. On Faust benchmark~\cite{FAUST}, the proposed UD$^2$E-Net outperforms state-of-the-art unsupervised methods by 24$\%$$\sim$37$\%$ on Inter challenge, and meanwhile achieves the best on Intra challenge even comparing to supervised methods.

\section{Qualitative Results}\label{qualitative}
\begin{figure*}[htbp]
	\centering
	\includegraphics[width=1.\textwidth]{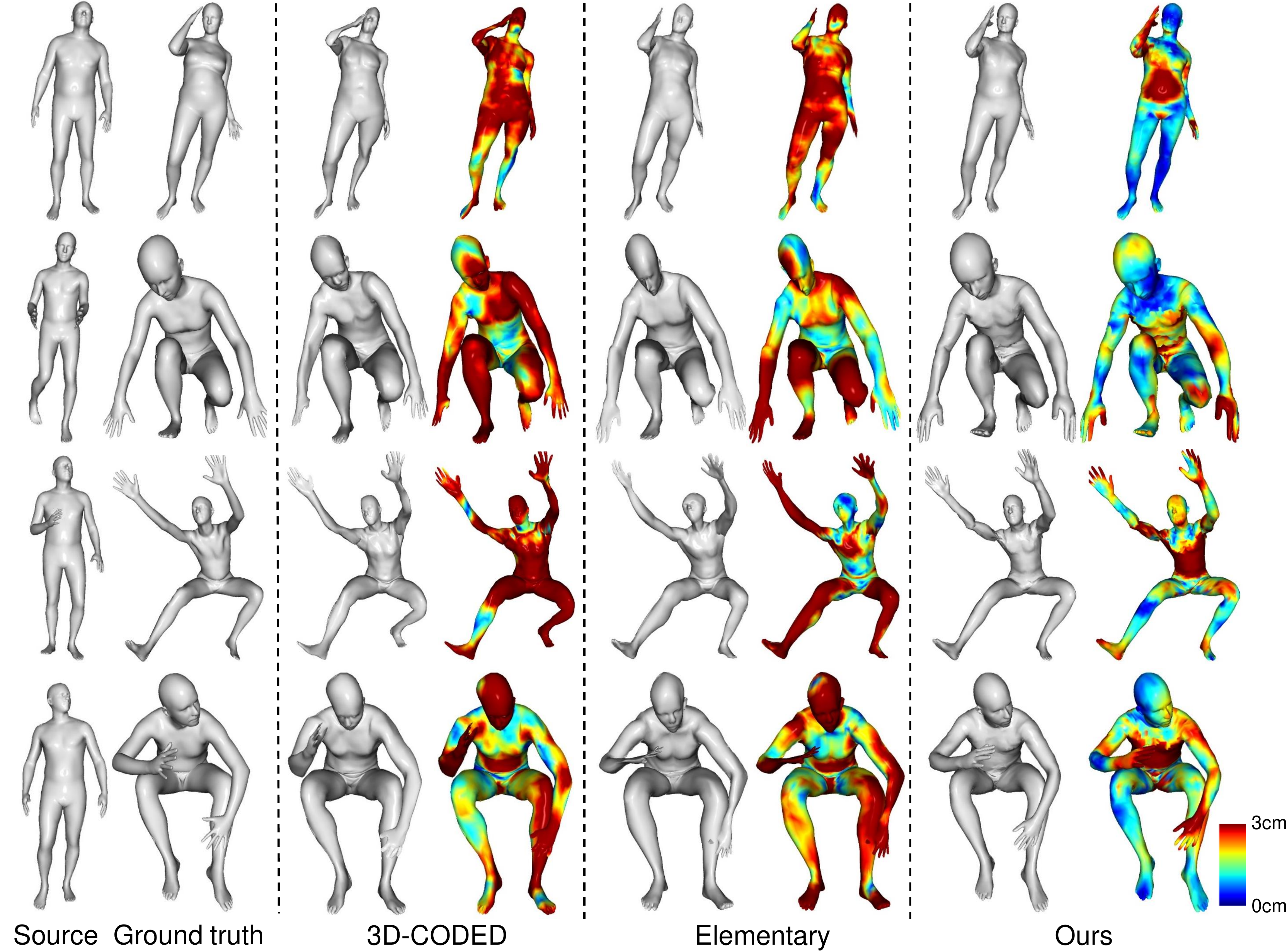}
	\caption{ Qualitative comparison on SURREAL 23k~\cite{surreal} with 3D-CODED~\cite{3d-coded} and Elementary~\cite{deprelle2019learning}. }\label{fig:comparison}
\end{figure*}

In Figure~\ref{fig:comparison}, we show more qualitative results for deformation prediction on SURREAL 23k~\cite{surreal} dataset, where comparing methods suffer from both unrealistic artifacts and large reconstruction error due to large non-rigid deformations, whereas our proposed UD$^2$E-Net can yield more natural and accurate deformations.

\section{Applications}\label{application}
\begin{figure*}[htbp]
	\centering
	\includegraphics[width=0.96\textwidth]{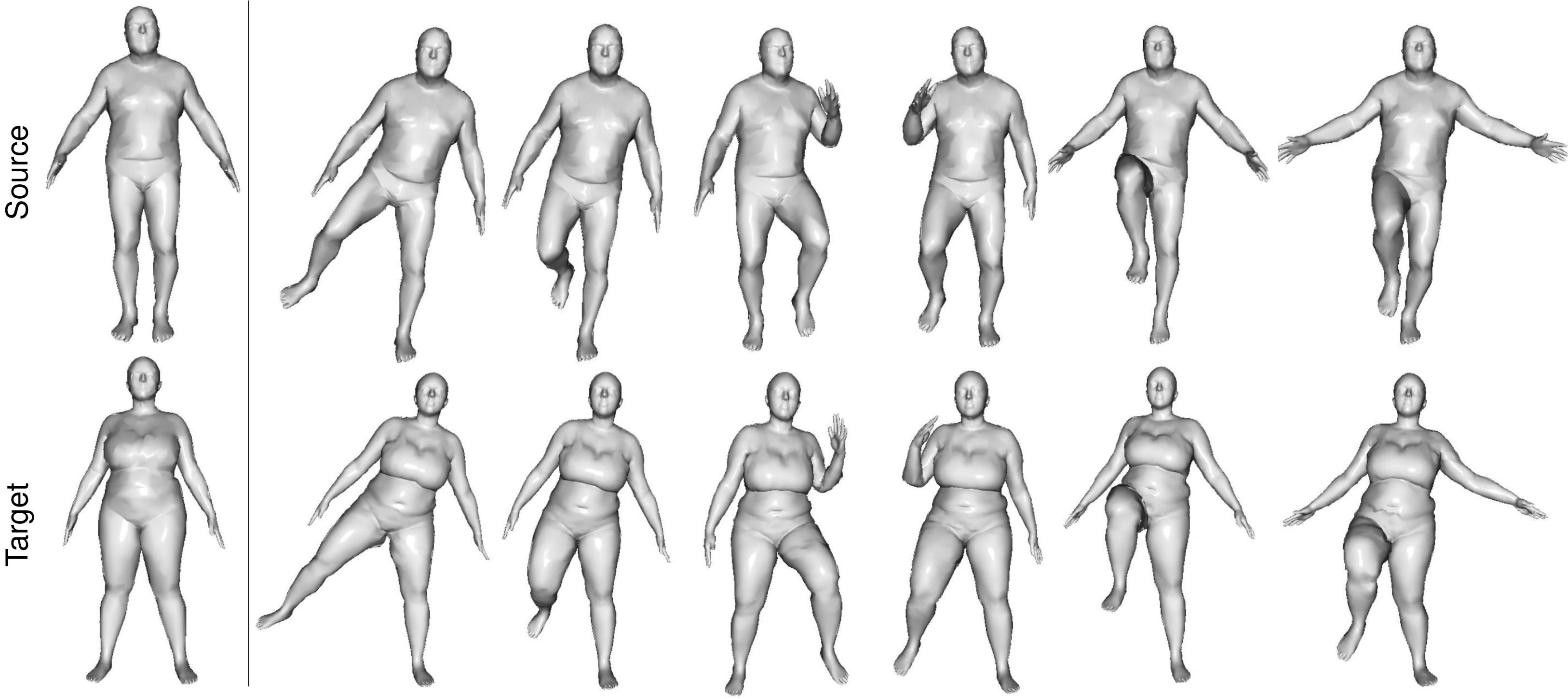}
	\caption{Human pose transfer results. The first column shows $A_0$ and $B_0$.}\label{fig:pose}
\end{figure*}

In Figure~\ref{fig:pose}, we show more human pose transfer results on DFAUST dataset~\cite{dfaust}, where the poses are successfully transferred from the source shape to the target shape. In Figure~\ref{fig:retrieval}, we show more shape retrieval results on SURREAL 23k dataset~\cite{surreal}, where the poses of the query and retrieved shapes are extremely similar.

\begin{figure*}[htbp]
	\centering
	\includegraphics[width=0.96\textwidth]{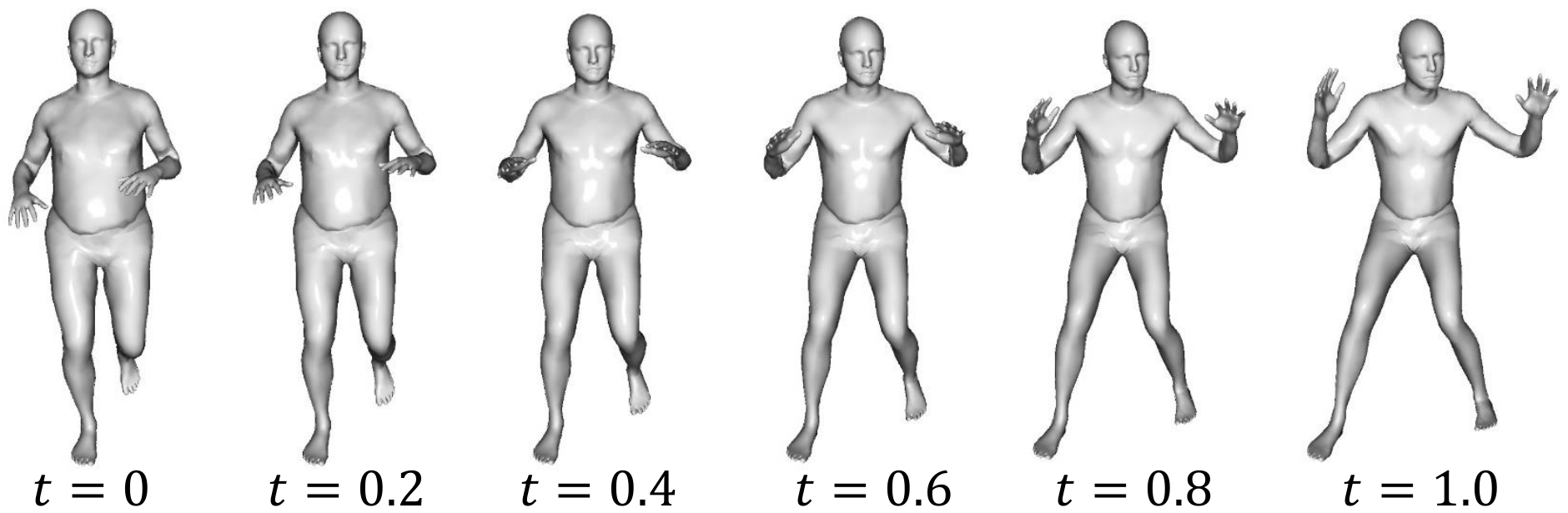}
	\caption{Shape interpolation results.}\label{fig:Arithmetic}
\end{figure*}

Here, we further evaluate UD$^2$E-Net on shape interpolation task. Although UD$^2$E-Net does not follow a strict Autoencoder architecture, which is known to be good at forming a latent space, it turns out that UD$^2$E-Net still forms a surprisingly well-behaved latent space. Given two target meshes, we linearly interpolate their global features $\mathbf{h}_0,\mathbf{h}_1$ by $\mathbf{h}_{inter}(t)=(1-t)\mathbf{h}_0+t\mathbf{h}_1$. As shown in Figure~\ref{fig:Arithmetic},  $\mathbf{h}_{inter}$ can yield plausible in-between meshes in both figure and pose.

\begin{figure*}[htbp]
	\centering
	\includegraphics[width=0.96\textwidth]{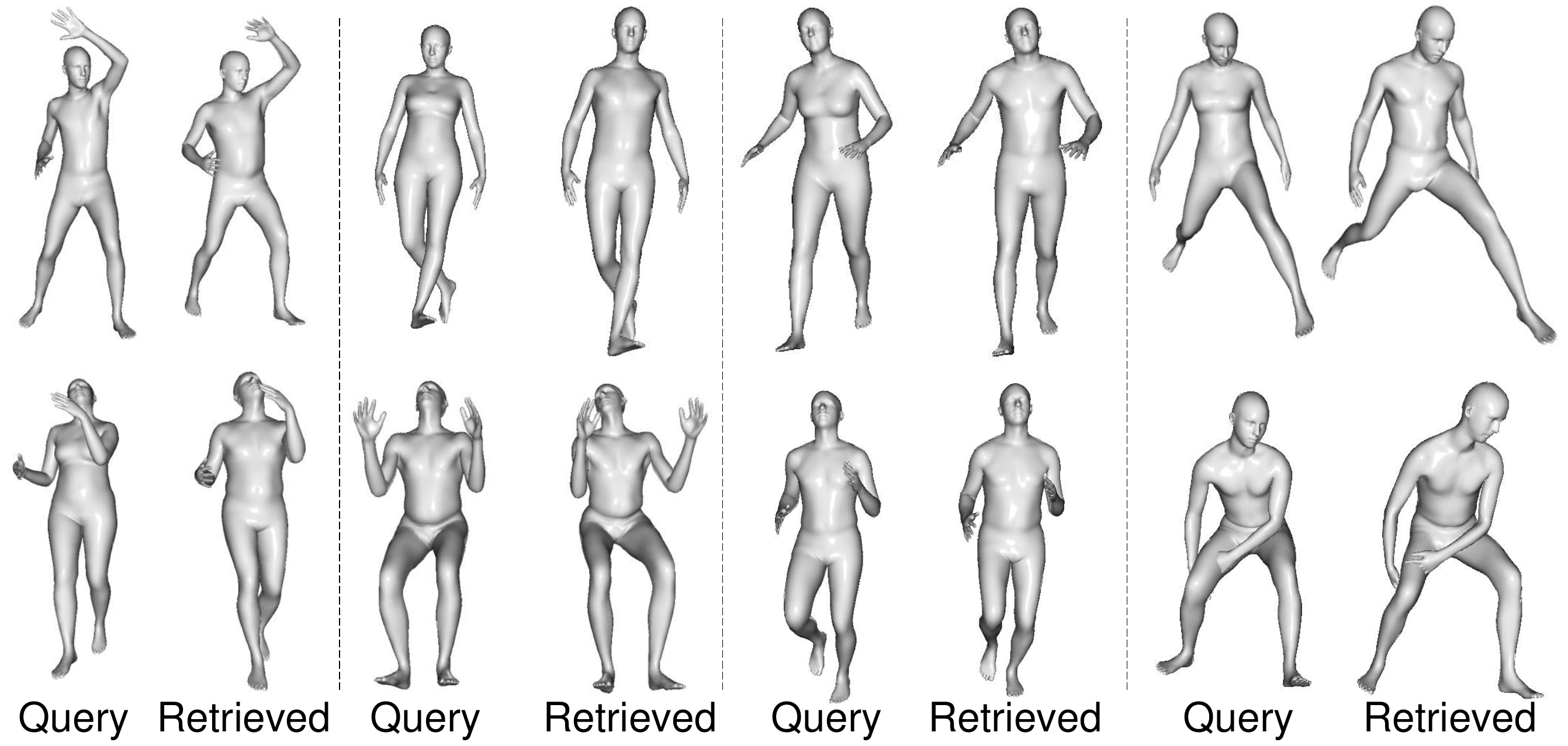}
	\caption{Results of shape retrieval on SURREAL 23k~\cite{surreal}, where the
		query shapes are on the left and retrieved shapes are on the right.}\label{fig:retrieval}
\end{figure*}

\end{appendices}

\end{document}